\documentclass{article}

\usepackage{arxiv}

\usepackage[utf8]{inputenc} 
\usepackage[T1]{fontenc}    
\usepackage{hyperref}       
\usepackage{url}            
\usepackage{booktabs}       
\usepackage{amsfonts}       
\usepackage{nicefrac}       
\usepackage{microtype}      
\usepackage{lipsum}		
\usepackage{graphicx}
\usepackage{natbib}
\usepackage{doi}
\usepackage{xcolor}
\usepackage{amsmath}
\usepackage{tabularx}
\usepackage{makecell}
\usepackage{tcolorbox}
\usepackage{mathtools} 
\usepackage{amssymb}
\usepackage{multirow}
\usepackage{subcaption}

\definecolor{bestgreen}{RGB}{0,120,70}
\definecolor{secondgreen}{RGB}{75,155,105}

\newcommand{\best}[1]{\textbf{\textcolor{bestgreen}{#1}}}
\newcommand{\second}[1]{\textcolor{secondgreen}{#1}}

\newtheorem{assumption}{Assumption}
\newtheorem{proposition}{Proposition}

\title{Evaluating Dynamical Fidelity through Predictive Structure in Physical Representations}

\author{
\href{https://orcid.org/0009-0003-5439-0139}{\includegraphics[scale=0.06]{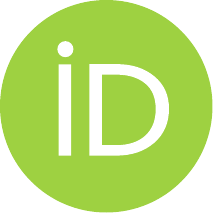}\hspace{1mm}Oskar Bohn Lassen}\\
Department of Technology,\\ Management, and Economics\\
Technical University of Denmark\\
\texttt{obola@dtu.dk}
\And
\href{https://orcid.org/0000-0002-2242-5056}{\includegraphics[scale=0.06]{orcid.pdf}\hspace{1mm}João Paulo de Souza Böger}\\
Department of Technology,\\ Management, and Economics\\
Technical University of Denmark\\
\texttt{jpade@dtu.dk}
\And
\href{https://orcid.org/0000-0001-5080-1234}{\includegraphics[scale=0.06]{orcid.pdf}\hspace{1mm}Simon Driscoll}\\
Department of Applied Mathematics\\ and Theoretical Physics\\
University of Cambridge\\
\texttt{sd2136@cam.ac.uk}
\And
\href{https://orcid.org/0000-0002-4775-3259}{\includegraphics[scale=0.06]{orcid.pdf}\hspace{1mm}Stephen I. Thomson}\\
Department of Mathematics\\ and Statistics\\
University of Exeter\\
\texttt{s.i.thomson@exeter.ac.uk}
\And
\href{https://orcid.org/0000-0002-1601-5683}{\includegraphics[scale=0.06]{orcid.pdf}\hspace{1mm}Sebastian Schemm}\\
Department of Applied Mathematics\\ and Theoretical Physics\\
University of Cambridge\\
\texttt{ss3299@cam.ac.uk}
\And
\href{https://orcid.org/0000-0001-6979-6498}{\includegraphics[scale=0.06]{orcid.pdf}\hspace{1mm}Filipe Rodrigues}\\
Department of Technology, \\Management, and Economics\\
Technical University of Denmark\\
\texttt{rodr@dtu.dk}
\And
\href{https://orcid.org/0000-0001-5457-9909}{\includegraphics[scale=0.06]{orcid.pdf}\hspace{1mm}Francisco C. Pereira}\\
Department of Technology, \\Management, and Economics\\
Technical University of Denmark\\
\texttt{camara@dtu.dk}
}

\renewcommand{\shorttitle}{}

\begin{document}
\maketitle

\begin{abstract}
Machine-learning models for physical systems are currently evaluated primarily through errors between predicted and reference states and, increasingly, through tests of physical consistency.
These metrics assess whether predictions are accurate and satisfy selected physical requirements, but provide limited insight into whether learned trajectories reproduce the underlying dynamics.
Domain experts examine such relationships through physical representations that expose relevant processes, interactions, and responses, but these analyses are often separated from typical machine-learning evaluation.
We introduce a practical framework for evaluating dynamical fidelity through predictive structure in physical representation spaces.
Experts define the representations, while reference trajectories determine which relationships are predictive and retained as evaluation tests.
We demonstrate the approach in atmospheric forecasting using ERA5 representations of planetary-wave activity and Northern Annular Mode evolution, and evaluate Pangu-Weather, GraphCast, and FengWu. 
The models exhibit distinct departures from reference predictive structure that are not reflected by conventional forecast errors.
The framework thereby turns domain-expert representations into systematic tests of learned physical dynamics without prescribing the relationships in advance.
\end{abstract}


\section{Introduction}
\label{sec:intro}
Machine-learning (ML) models are increasingly used to predict the evolution of complex physical systems across materials science, molecular dynamics, and climate science~\citep{noe_machine_2020,pfaff_learning_2021,reichstein_deep_2019,batzner_e3-equivariant_2022}. 
These domains are also described by decades of physical theory and numerical modelling, giving us substantial prior knowledge of how the dynamics should behave.
While physical constraints and inductive biases can be incorporated into such models~\citep{karniadakis_physics-informed_2021}, their success has been mixed, and many leading models remain highly flexible and largely data-driven. 
This creates a need to use physical knowledge not only to shape models, but to assess whether their predictions reproduce known dynamics.

Evaluation of these models remains dominated by errors between predicted and reference states, which quantify whether a model reaches the correct state but say little about dynamical fidelity: whether it follows the correct dynamical evolution.
Recent work therefore complements predictive accuracy with tests of physical consistency, conservation laws, symmetries, and balance relationships~\citep{hansen_learning_2023,liu_harnessing_2024}. 
In principle, a complete set of physical constraints, sufficient to characterize the underlying dynamics, would also establish complete dynamical fidelity. 
In practice, however, only a limited subset of such constraints can typically be specified and evaluated, so physical consistency restricts the class of admissible dynamics without establishing that the learned evolution matches the underlying system.
Across the physical sciences, model validation has therefore also long relied on domain-specific diagnostics of dynamical behaviour, from energy-transfer statistics in turbulence to process-oriented diagnostics in climate models~\citep{ortali_numerical_2022,mohanty_evaluating_2023,maloney_process-oriented_2019,nowack_causal_2020,lassen_investigating_2026,wu_assessing_2026}. 
These diagnostics probe whether particular processes, interactions, or response relationships realized by the reference system are preserved by the model. 
Yet they are usually developed within individual domains and evaluated separately from standard ML metrics, decoupling process-level assessment from routine ML evaluation and making it difficult to determine whether gains in predictive accuracy correspond to more faithful dynamics.

We operationalize process-level dynamical fidelity by identifying predictive relationships in reference trajectories within expert-defined physical representations and testing whether learned models preserve them.
These representations define a restricted physical description of the system and therefore another partial view of its dynamics, complementary to testing a finite set of physical-consistency constraints.
We assume only that this representation space contains physically relevant sources of predictability for the process under investigation, not that it characterizes the full dynamics.
Rather than prescribing which relationships a learned model should satisfy, we identify predictive relationships directly from the reference trajectories within this space, freeze them, and test whether model-generated trajectories preserve the same conditional structure.
Experts thereby determine the physical vocabulary in which dynamical behaviour is expressed, while the reference trajectories determine which relationships within that vocabulary are actually predictive.

We instantiate this framework in machine-learning weather forecasting, where models such as Pangu-Weather, GraphCast, and FengWu now produce skillful global forecasts at a fraction of the cost of conventional numerical weather prediction~\citep{bi_accurate_2023,lam_learning_2023,chen_operational_2025}. 
Evaluation remains largely state-based through benchmarks such as WeatherBench 2~\citep{rasp_weatherbench_2024}, while PhysMetrics.Weather extends this evaluation to conservation laws, spectra, and balance relationships~\citep{kasteleyn_physmetricsweather_2026}. 
At the same time, atmospheric science has a long tradition of process-oriented diagnostics for evaluating wave propagation, forcing, and circulation response~\citep{maloney_process-oriented_2019, kim_process-oriented_2014, edmon_eliassen-palm_1980}. 
This combination makes weather forecasting a natural test bed for evaluating dynamical fidelity beyond state-space accuracy and physical consistency alone.
We focus on stratospheric wave--mean-flow dynamics and the subsequent evolution of the Northern Annular Mode (NAM). 
Using the ERA5 atmospheric reanalysis as the reference system, we construct an expert-defined representation space spanning planetary-wave activity, vertical propagation, wave forcing, background circulation, and circulation response. 
Within this space, we identify predictive relationships in ERA5 and test whether ML weather models preserve their conditional consequences, beyond reproducing the atmospheric state itself.

Our contributions are:
\begin{itemize}
\item We formalize state-space accuracy, physical consistency, and dynamical fidelity, show that physical consistency does not imply dynamical fidelity, and introduce a framework that discovers process-level dynamical fidelity in expert-defined representations.
\item We instantiate the framework for stratospheric wave--mean-flow dynamics and NAM evolution, using ERA5 to discover robust conditional relationships and show that process-level dynamical fidelity reveals model differences across GraphCast, FengWu, and Pangu-Weather not captured by conventional forecast error.
\end{itemize}

\section{From Physical Consistency to Dynamical Fidelity}
\label{sec:dynamical-fidelity}

Let \(x_t \in \mathcal{X}\) denote the state of a physical dynamical system at time \(t\). 
A machine-learning model defines a learned state-transition operator
\(
\widehat{x}_{t+\Delta t}
=
f_{\theta}(\widehat{x}_t),
\)
where \(f_{\theta}\) is applied repeatedly to produce an autoregressive trajectory. Predictions are initialised from a reference state \( x_{t_0}  = \widehat{x}_{t_0} \), and subsequent hatted states are generated by the model.
We use a one-step autoregressive model for notational simplicity, but the formalization below applies to models conditioned on multiple preceding states or predicting multiple future states jointly.

\paragraph{Prediction from data.}
Machine-learning models of physical dynamical systems are commonly trained by minimising prediction error over observed state transitions,
\[
\theta^\star
=
\arg\min_{\theta}
\;
\mathbb{E}_{(x_t,x_{t+\Delta t})\sim p_{\mathrm{train}}}
\left[
\ell\!\left(f_\theta(x_t),x_{t+\Delta t}\right)
\right],
\]
where \(p_{\mathrm{train}}\) is the training distribution and \(\ell\) measures discrepancy between predicted and reference states, possibly with added regularisation.
Such training rewards any relationship that improves prediction within \(p_{\mathrm{train}}\), whether it reflects a known physical mechanism, a previously unresolved regularity, or a statistical shortcut~\citep{geirhos_shortcut_2020}.
This flexibility is a strength, the learned dynamics are not restricted to what a numerical scheme already encodes, but low prediction error does not identify which relationships were learned, and two models can perform similarly while relying on different ones~\citep{damour_underspecification_2022}.
If those behave differently outside \(p_{\mathrm{train}}\) or over long rollouts, errors accumulate and trajectories diverge~\citep{sun_can_2025,sanchez-gonzalez_learning_2020}.

\paragraph{Physical consistency.}
Physical-consistency diagnostics assess whether learned dynamics satisfy known requirements of the underlying physical system. 
Such requirements may also be imposed during training through constrained architectures, physical losses, or hybrid models, but here we consider them purely as evaluation criteria~\citep{karniadakis_physics-informed_2021,greydanus_hamiltonian_2019, batzner_e3-equivariant_2022,hansen_learning_2023,kochkov_neural_2024}; (Appendix~\ref{app:hybrid}).
To rigorously examine the limitations of physical consistency evaluations, consider a dynamical system
\[
    \dot{x}(t)=F(x(t)),
\]
where $F$ denotes the vector field mapping each state to its instantaneous time derivative, and let $\mathfrak{F}$ denote the hypothesis class. 
A sufficiently rich collection of physical relations can itself be viewed as a characterization of the dynamics~\citep{landau_chapter_1976}. 
Formally, let
\[
    \mathcal{Q} :
    \mathfrak{F}\rightarrow\mathcal{Y},
    \qquad
    F\mapsto
    \mathcal{Q}(F)
    =
    \bigl(Q_\alpha[F]\bigr)_{\alpha\in\mathcal{I}}
\]
where $\alpha$ indexes individual physical relations, $\mathcal{I}$ is the index set of the complete collection, and $\mathcal{Y}$ denotes the space of all possible collections of values taken by these physical relations. 
Thus, $\mathcal{Q}(F)$ represents the full set of physical constraints associated with the dynamics defined by $F$, including governing-equation relations, conservation laws, symmetries, balance relations, constitutive constraints, or other physical requirements.

In the complete case, the collection of physical relations $\mathcal Q(F)$ separates admissible dynamics,
\[
    \mathcal{Q}(F_1)=\mathcal{Q}(F_2)
    \quad\Longrightarrow\quad
    F_1=F_2.
\]
Thus, when $\mathcal{Q}$ is sufficiently rich to be injective, the complete physical characterization $\mathcal{Q}(F)$ and the vector field $F$ contain equivalent information about the dynamics.

Physical-consistency evaluation generally has access to only a subset of this characterization. 
Let $J\subset\mathcal{I}$ index the physical relations that are known and practically testable. 
We write each $Q_j$ as a residual such that
\[
    Q_j[F]=0
\]
when the underlying dynamics satisfy the corresponding physical relation. 
The same requirement can then be evaluated for any candidate vector field.

Let $\tilde{F}\in\mathfrak{F}$ denote an arbitrary candidate vector field. 
The known constraints restrict the possible dynamics to
\[
    \mathfrak{F}_{J}
    =
    \left\{
        \tilde{F}\in\mathfrak{F}:
        Q_j[\tilde{F}]=0
        \ \mathrm{for}\ j \in J
    \right\}.
    \label{eq:physical_equivalence_class}
\]
Unless these constraints are sufficient to identify the dynamics, $\mathfrak{F}_{J}$ may contain many vector fields $\tilde{F}\neq F$. 
Adding another valid physical constraint $j^\ast\notin J$ can only further restrict this set, such that
\[
    F\in\mathfrak{F}_{J\cup\{j^\ast\}}
    \subseteq
    \mathfrak{F}_{J}.
\]
However, this restriction alone provides no guarantee that the remaining admissible vector fields are close to the true dynamics. 
For a chosen norm on vector fields, the identifiability radius is defined as
\[
R_J(F)
=
\sup_{\tilde{F}\in\mathfrak{F}_{J}}
\lVert \tilde{F}-F\rVert , \quad R_J(F) \in [0, \infty],
\]
which satisfies $R_{J'}(F)\leq R_J(F)$ whenever $J\subseteq J'$, but need not be small for any $J\subsetneq\mathcal I$ that does not identify $F$.

Empirical evaluation introduces a further loss of information.
Even when a physical relation is known exactly, it can generally be evaluated only at finite spatial and temporal resolution. 
For example, the continuous relation $\dot{x}(t)=F(x(t))$ can only be tested through a temporal discretization such as
\[
    \frac{x_{t+\Delta t}-x_t}{\Delta t}
    \approx F(x_t),
\]
together with an analogous discretization in space. 
Thus, even the known constraints in $\mathcal{Q}_{J}$ are generally tested only through spatiotemporally coarsened approximations of their continuous counterparts.
A learned system may consequently satisfy every evaluated conservation law, symmetry, or balance relation while still realizing different transport, propagation, interaction, or response behaviour~\citep{bonavita_limitations_2024}.
Appendix~\ref{app:coarsening} further formalizes and proves the complementarity between physical-consistency and process-level evaluation introduced next.

\paragraph{Process-level evaluation of dynamical fidelity.}
Because complete dynamical fidelity is generally not directly testable, it can instead be assessed through process-level diagnostics that expose particular aspects of the system's evolution. 
Let
\[
\mathcal{P}_k\!\left(x_{t:t+L}\right),
\qquad k=1,\ldots,K,
\]
denote the $k$-th process diagnostic applied to a trajectory of length $L$.
Such diagnostics are well established across the physical sciences and are typically specified individually for a particular process~\citep{ortali_numerical_2022,mohanty_evaluating_2023,maloney_process-oriented_2019}.
Depending on the application, $\mathcal{P}_k$ may characterize a flux, transport, propagation, forcing, or response, returning a scalar, a field, or a full space--time diagnostic.
Agreement in a given diagnostic can be measured as
\[
\mathcal{E}^{\mathrm{dyn}}_k
=
d_k\!\left(
\mathcal{P}_k(\widehat{x}_{t:t+L}),
\mathcal{P}_k(x_{t:t+L})
\right),
\]
where $d_k$ compares predicted and reference diagnostics. 
Small $\mathcal{E}^{\mathrm{dyn}}_k$ therefore indicates agreement in the particular aspect of the dynamics exposed by $\mathcal{P}_k$. 

\section{Evaluating dynamical fidelity through expert-defined representations and relationship discovery}
\label{sec:expert-representations}
Process-level evaluation in the physical sciences has traditionally relied on diagnostics specified individually for a particular process, as represented by $\mathcal P_k$ above.
Designing and interpreting such tests requires  substantial domain expertise and is therefore often separated from the standard machine-learning evaluation pipeline, with important model deficiencies becoming apparent only through subsequent domain-specific analysis. 
To make process-level evaluation more systematic and easier to integrate into model development, we separate the definition of the physical representation space from the relationships evaluated within it. 
This framework proposes that experts define a physically meaningful representation space, while the predictive relationships to be evaluated are discovered from reference trajectories and subsequently frozen for model evaluation.

\subsection{Expert-defined physical representations}
\label{sec:expert_defined_method}

The expert-defined representations consist of physically meaningful transformations of the state, chosen to expose relevant sources, transports, interactions, and responses.
Their spatial and temporal evolution provides the physical vocabulary within which predictive relationships are subsequently discovered.
We make this assumption explicit:
\begin{assumption}[Physically informative representation]
\label{ass:expert-representation}
The expert-defined representation space contains physically relevant sources of predictability for the processes under investigation.
\end{assumption}
This does not require the representation space to be sufficient for the complete system dynamics.
It only requires that some combinations of represented quantities, spatial regions, and preceding times contain information about the subsequent evolution of other physically relevant quantities.

Let $\mathcal X$ denote the physical state space, with $x_t\in\mathcal X$ denoting the system state at time $t$. We define $R$ expert-specified representations by
\[
\mathcal G=\{g_1,\ldots,g_R\},
\qquad
g_j:\mathcal X\rightarrow\mathcal Z_j,
\quad
j=1,\ldots,R,
\]
where $\mathcal Z_j$ is the representation space associated with $g_j$ and may be scalar-, vector-, or field-valued. The joint representation map is defined as
\[
G:\mathcal X\rightarrow
\mathcal Z_1\times\cdots\times\mathcal Z_R,
\qquad
G(x_t)
=
\bigl(
g_1(x_t),\ldots,g_R(x_t)
\bigr).
\]
For each representation $g_j$, let $A\in\mathcal A_j$ denote an admissible spatial region and let $\rho_A$ denote a corresponding spatial operator. We then define the spatially aggregated representation as
\(
g_j^A(x_t)
=
\rho_A\!\left(g_j(x_t)\right).
\)

\subsection{Discovering predictive relationships and evaluating dynamical fidelity}
\label{sec:identify_relationships_method}

Let \(t_c\) denote a cut time separating conditioning history from the future trajectory of interest.
Let $W\in\mathcal W$ denote a temporal window relative to $t_c$, with $W\subseteq(-\infty,0]$, and let $\psi\in\Psi$ denote an admissible temporal aggregation operator.

A candidate source variable is specified by a representation $j$, spatial region $A$, temporal window $W$, and operator $\psi$
\[
X_{t_c}^{j,A,W,\psi}
=
\psi\left(
\left(
g_j^A(x_{t_c+\tau})
\right)_{\tau\in W}
\right).
\]
A conditioning vector consists of $M$ such source variables, which may span different representations, spatial regions, temporal windows, and aggregation operators
\[
\mathbf X_{t_c}
=
\left(
X_{t_c}^{j_1,A_1,W_1,\psi_1},
\ldots,
X_{t_c}^{j_M,A_M,W_M,\psi_M}
\right).
\]
For a target representation \(k\) and spatial region \(B\in\mathcal A_k\), its future trajectory after the cut time \(t_c\) is defined as
\[
\mathbf Y_{>t_c}^{k,B}
=
\left(
g_k^B(x_{t_c+\tau_1}),
\ldots,
g_k^B(x_{t_c+\tau_K})
\right),
\qquad
0<\tau_1<\cdots<\tau_K.
\]

The relationship-discovery problem is to identify source--target pairs for which the source strongly conditions the target distribution in the reference trajectories.
Let $\mathcal C$ denote the admissible catalogue of candidate pairs $r=(\mathbf X_{t_c}^{(r)},\mathbf Y_{>t_c}^{(r)})$.
Our method evaluates these candidates on reference trajectories with a discovery and validation procedure $\mathcal D_{\mathrm{ref}}$ that maps the catalogue to a retained set,
\[
\mathcal R_{\mathrm{ref}}^\star
=
\mathcal D_{\mathrm{ref}}(\mathcal C),
\qquad
\mathcal R_{\mathrm{ref}}^\star\subseteq\mathcal C.
\]
The procedure $\mathcal D_{\mathrm{ref}}$ may be model-based or model-free, causal or purely predictive, and may search candidates individually, greedily, or jointly.
The candidate space can grow rapidly if all admissible representations are considered as targets and paired with all possible sources. 
Established physical theory, prior literature, and literature-informed AI agents can therefore propose and refine \(\mathcal C\), while reference trajectories determine which relationships are retained. 
Once discovered, the retained relationships are frozen before model evaluation and remain expressed in the expert-defined physical representation space.

For each statistically validated relationship \(r\in\mathcal R_{\mathrm{ref}}^\star\), we consider both the joint conditional trajectory distribution and its lead-time marginals,
\[
p_{\mathrm{ref}}
\left(
\mathbf Y_{>t_c}^{k,B}
\mid
\mathbf X_{t_c}
\right),
\qquad
p_{\mathrm{ref}}
\left(
g_k^B(x_{t_c+\tau_\ell})
\mid
\mathbf X_{t_c}
\right),
\quad
\ell=1,\ldots,K.
\]
We can then evaluate ML model performance by quantifying agreement with the reference conditional relationship using an application-appropriate discrepancy
\[
\mathcal{E}^{\mathrm{dyn}}_{k,B}
=
S\Bigl(
p_{\mathrm{ref}}\bigl(\mathbf Y^{k,B}_{>t_c}\mid\mathbf X_{t_c}\bigr),
p_{\mathrm{model}}\bigl(\widehat{\mathbf Y}^{k,B}_{>t_c}\mid\mathbf X_{t_c}\bigr)
\Bigr),
\]
where \(S\) may compare the full distributions or selected statistics such as their conditional means and variances. 
Each retained relationship thus acts as a process diagnostic $\mathcal P_k$ in the sense of Section~\ref{sec:dynamical-fidelity}, discovered from the reference rather than specified in advance, and can therefore detect errors that invariants and budgets miss (Appendix~\ref{app:coarsening}).

\section{Instantiating Dynamical Fidelity in Stratospheric Circulation}
\label{sec:instantiating-expert-representations}
We instantiate the framework in atmospheric forecasting, focusing on the Northern Hemisphere stratospheric circulation.
This setting is particularly suitable because established diagnostics of planetary-wave activity and wave--mean-flow interaction provide an expert-defined representation space for predicting subsequent Northern Annular Mode (NAM) evolution.

\subsection{Scientific target: Northern Annular Mode evolution}
\label{sec:nam_target}
We use the Northern Annular Mode (NAM) as a continuous measure of the large-scale circulation and stratosphere--troposphere coupling~\citep{baldwin_stratospheric_2001,baldwin_sudden_2021}.
Stratospheric NAM anomalies can persist and influence the tropospheric circulation for several weeks~\citep{domeisen_role_2020,baldwin_sudden_2021}, making their early evolution relevant for longer-range prediction.

At pressure level $p$ and time $t$, we define the NAM index $N(p,t)$ as the standardized projection of the geopotential-height anomaly $Z'(\lambda,\phi,p,t)$ north of $20^\circ$N onto its leading empirical orthogonal function (EOF) $e_p(\lambda,\phi)$,
\[
N(p,t)
=
\frac{c(p,t)-\mu_{c,p}}{\sigma_{c,p}},
\qquad
c(p,t)
=
\left\langle
\sqrt{\cos\phi}\, Z'(\cdot,\cdot,p,t),
e_p
\right\rangle,
\]
where $c(p,t)$ is the area-weighted EOF projection coefficient, $\lambda$ and $\phi$ denote longitude and latitude, $\langle\cdot,\cdot\rangle$ denotes the inner product over the latitude--longitude grid, and $\mu_{c,p}$ and $\sigma_{c,p}$ are the reference mean and standard deviation of $c$ at pressure level $p$.

We focus on the NAM averaged over the 50--100\,hPa layer, denoted $N_{50:100}(t)$, and restrict the reference cohort to approximately neutral initial states,
\[
\mathcal T
=
\left\{
t_c:
|N_{50:100}(t_c)|\leq 0.3
\right\}.
\]
Across winters 1979/80--2024/25 in the ERA5 global atmospheric reanalysis, we obtain $n=772$ admissible initialization times.
Conditioning on a comparable initial NAM state shifts the question from persistence of the circulation itself to which preceding physical conditions distinguish its subsequent evolution.
In this application, all candidate relationships $r\in\mathcal C$ share the same target, hence, for each initialization $t_c\in\mathcal T$,
\[
\mathbf Y_{>t_c}^{(r)}
=
\mathbf Y_{>t_c}
=
\left(
N_{50:100}(t_c+1),
\ldots,
N_{50:100}(t_c+10)
\right),
\qquad r\in\mathcal C.
\]
Details of the reference climatology, EOF fitting, and normalization are given in Appendix~\ref{app:nam-convention}.

\subsection{Physical representations and candidate precursor space}

\begin{figure}[t]
\centering
\includegraphics[width=\textwidth]{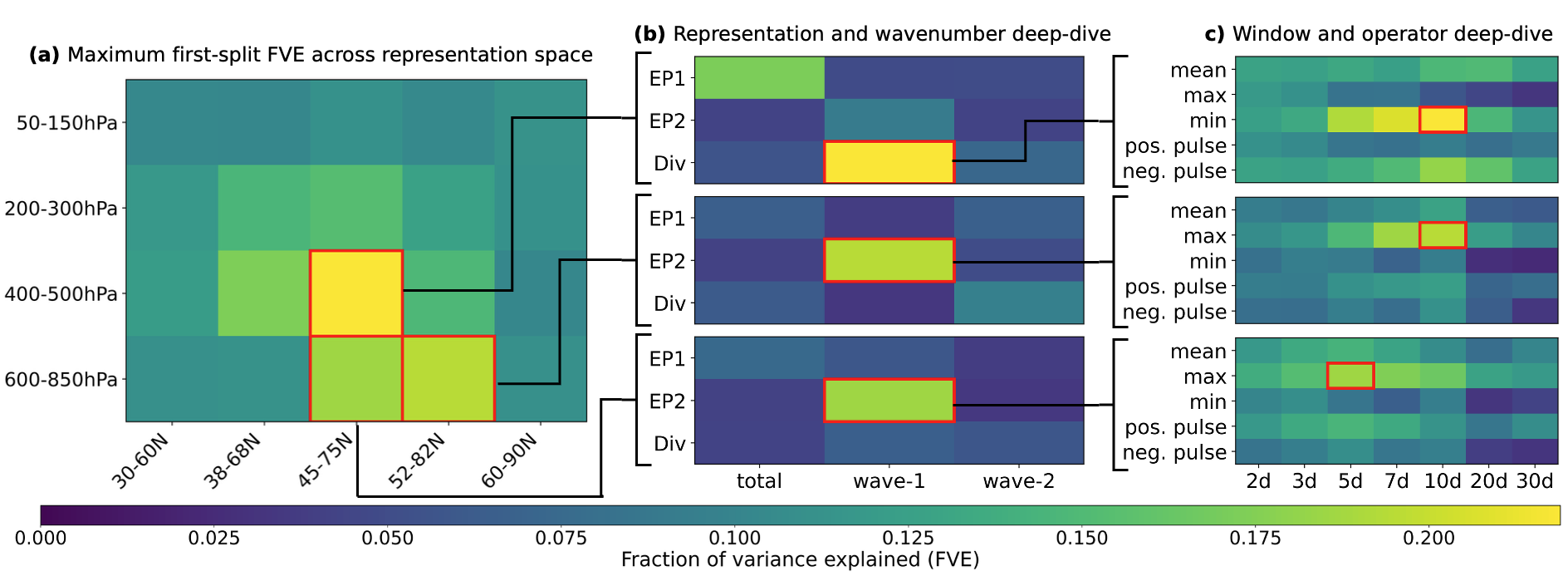}
\caption{
First-level predictive structure in ERA5 before robustness and permutation-FDR filtering.
\textbf{(a)} Maximum FVE across the candidate space by latitude--pressure region.
\textbf{(b)} Representation/wavenumber deep-dive for the three highest-FVE regions.
\textbf{(c)} Window/operator deep-dive for their strongest configurations.
Outlined cells indicate the successive maxima.
}
\label{fig:predictive-landscape}
\end{figure}

Having defined future NAM evolution as the target, we construct a physically meaningful space of candidate precursors from established diagnostics of planetary-wave propagation and wave--mean-flow interaction~\citep{edmon_eliassen-palm_1980,baldwin_stratospheric_2001}.
For $q\in\{u,v,T\}$, denoting zonal wind, meridional wind, and temperature, we decompose the flow into its zonal mean $\overline q(\phi,p,t)$ and eddy departure $q'=q-\overline q$, separating the background circulation from wave-related variability.
The eddy covariances \(\overline{v'T'}\) and \(\overline{u'v'}\) measure the meridional eddy transport of heat and zonal momentum, and enter the quasi-geostrophic Eliassen--Palm (EP) flux~\citep{edmon_eliassen-palm_1980},
\[
\mathbf F=(F_\phi,F_p),
\qquad
F_\phi\propto-\overline{u'v'},
\qquad
F_p\propto\overline{v'T'},
\]
whose components diagnose meridional and vertical propagation of wave activity, and whose divergence \(D\equiv\nabla\!\cdot\!\mathbf F\) diagnoses the resulting wave forcing of the zonal-mean circulation.
Our candidate representation space includes these quantities together with the \(m=1\) and \(m=2\) zonal-wavenumber contributions \(\mathbf F^{(m)}\) and \(D^{(m)}\), obtained by Fourier decomposition in longitude, which dominate stratospheric variability.
Together with the NAM defined above, the expert representation is
\[
G(x_t)
=
\left\{
\overline u,\;
\overline T,\;
F_\phi,\;
F_p,\;
D,\;
F_\phi^{(1)},F_p^{(1)},D^{(1)},\;
F_\phi^{(2)},F_p^{(2)},D^{(2)},\;
N
\right\}.
\]
These representations span the background circulation, planetary-wave propagation and forcing, and the resulting circulation response.

For discovery, the field representations $g_j$ are converted to climatologically standardized anomalies on the $5^\circ$ latitude bands and pressure levels before spatial aggregation into $g_j^A$, where $\rho_A$ denotes an unweighted mean over the $5^\circ$ latitude bands and pressure levels. A candidate source is defined as
\[
X_{t_c}^{j,A,W_d,\psi}
=
\psi\left(
\left(
g_j^A(x_{t_c+\tau})
\right)_{\tau\in W_d}
\right).
\]
where temporal operators $\psi\in\{\mathrm{mean},\mathrm{max},\mathrm{min}, \mathrm{positive\ impulse},\mathrm{negative\ impulse}\}$ are applied over trailing windows $W_d$, with \( d\in\{2,3,5,7,10,20,30\}\ \text{days}\) and \(W_d=\{-d,-d+6\,\mathrm{h},\ldots,-6\,\mathrm{h}\}\).
We index each admissible combination $(j,A,W_d,\psi)$ by $r$, yielding a scalar candidate source. Paired with the common future-NAM trajectory defined above, each scalar source defines a candidate relationship $r\in\mathcal C$.
Enumerating all candidate sources yields 7{,}875 candidate relationships.
Additional details are provided in Appendix~\ref{app:wave_theory}-\ref{app:relationship-search}.

\begin{figure}[t]
    \centering
    \begin{subfigure}[b]{0.32\textwidth}
        \centering
        \includegraphics[width=\textwidth]{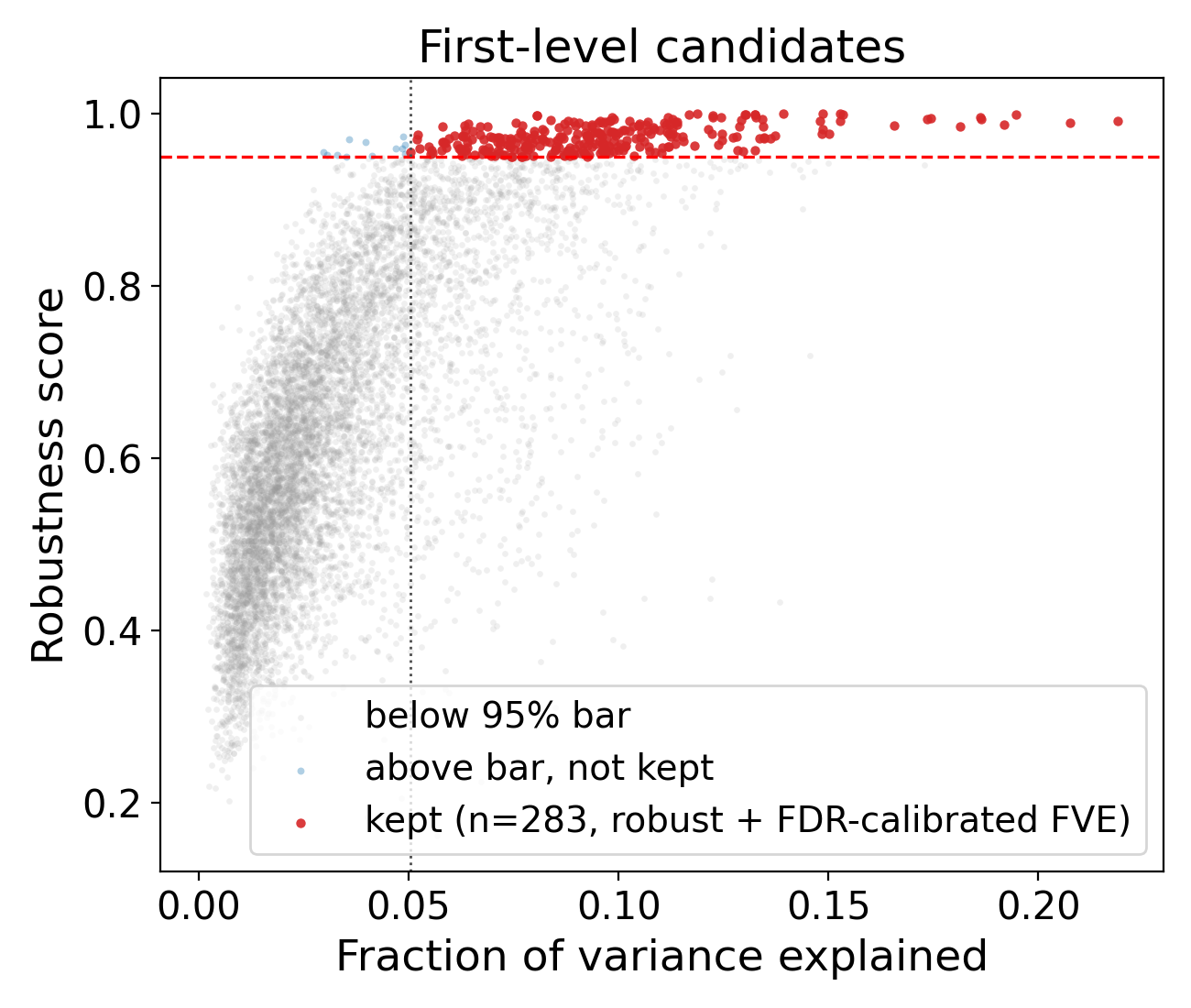}
        \caption{First-level candidates}
        \label{fig:selection-first}
    \end{subfigure}
    \hfill
    \begin{subfigure}[b]{0.32\textwidth}
        \centering
        \includegraphics[width=\textwidth]{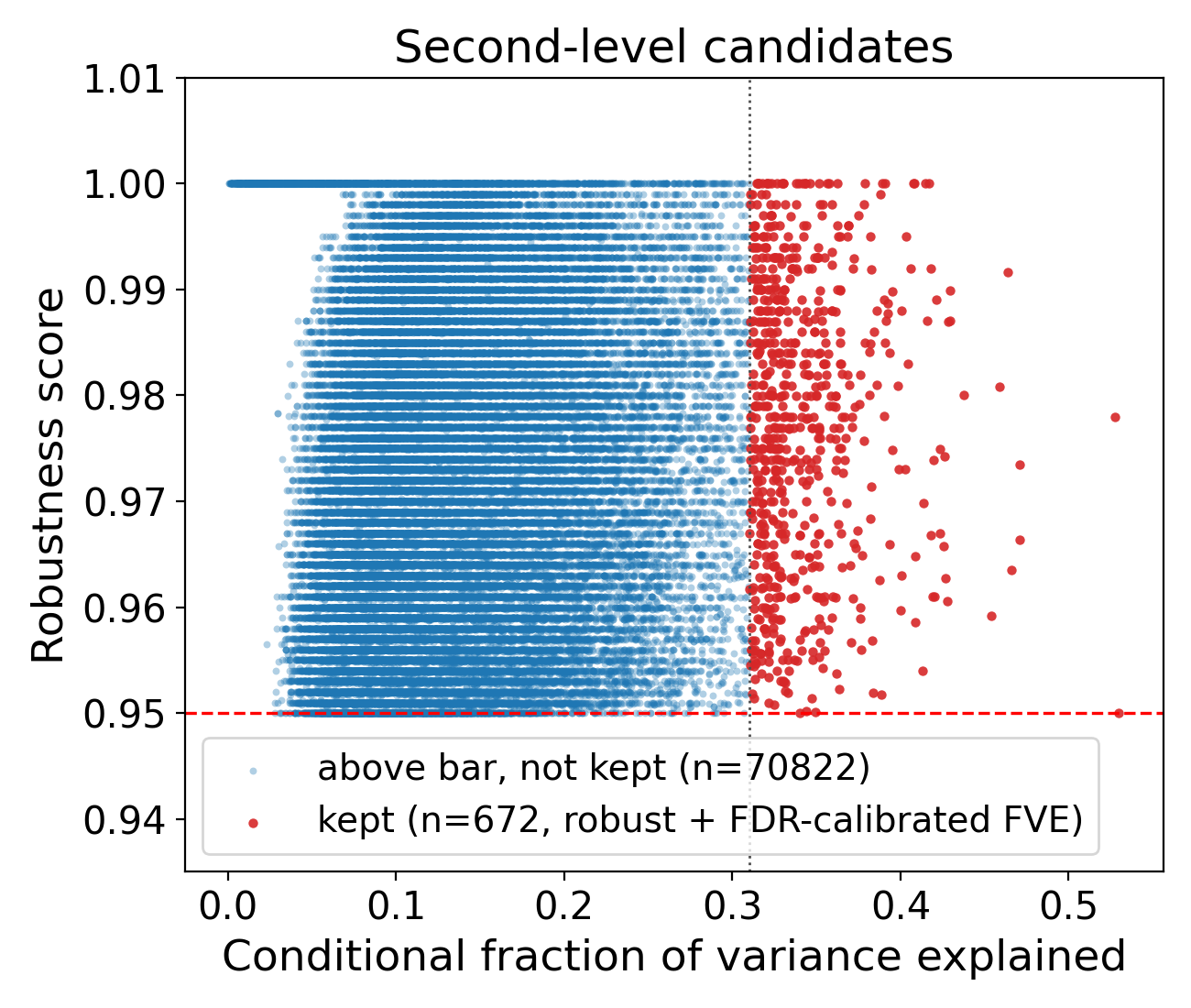}
        \caption{Second-level candidates}
        \label{fig:selection-second}
    \end{subfigure}
    \hfill
    \begin{subfigure}[b]{0.32\textwidth}
        \centering
        \includegraphics[width=\textwidth]{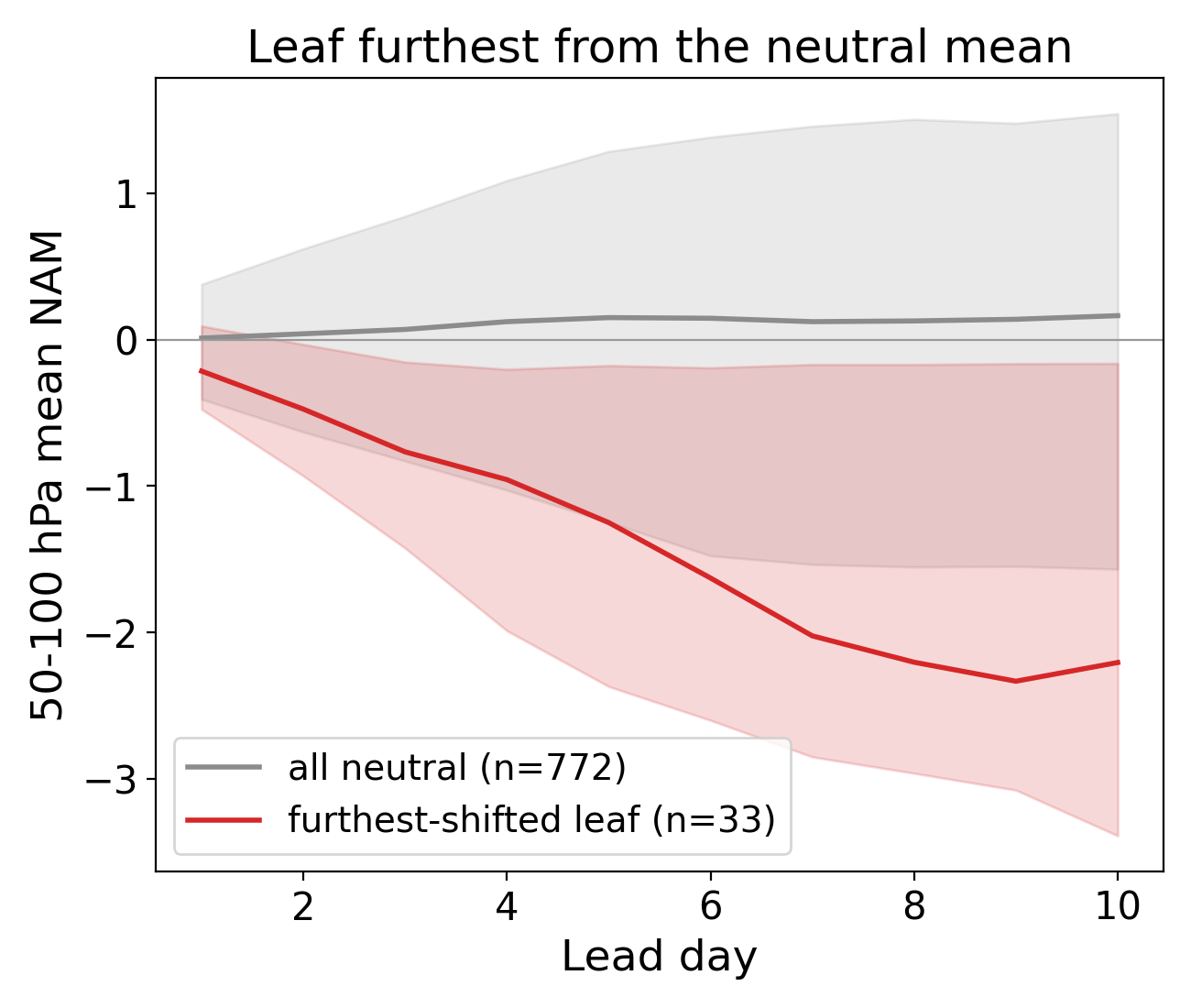}
        \caption{Example conditional trajectory}
        \label{fig:selection-best-leaf}
    \end{subfigure}

    \caption{%
        Selection of robust predictive relationships.
        \textbf{(a)} Of 7{,}875 first-level candidates, 295 satisfy the 95\% directional-stability criterion; permutation-FDR calibration retains 283 at FVE $\geq 0.0504$.
        \textbf{(b)} Second-level search within their frozen branches evaluates 4{,}358{,}681 conditional candidates, of which 71{,}494 satisfy 95\% stability and 672 remain after permutation-FDR calibration (conditional FVE $\geq 0.3102$).
        \textbf{(c)} Example of leaf that deviates from the neutral mean, isolating initially neutral states that evolve strongly. Lines show winter-weighted medians and shading 10--90\% ranges.
    }
    \label{fig:selection}
\end{figure}

\subsection{Relationship discovery and frozen model evaluation}
\label{sec:relationship-search}

Using the $n=772$ neutral-NAM initializations, we consider each candidate relationship $r\in\mathcal C$ and threshold split $X^{(r)}\leq c$ where each split partitions $\mathcal T$ into $\mathcal T_{0}$ and $\mathcal T_{1}$.
Separation is measured by the winter-weighted fraction of variance explained,
\[
\mathrm{FVE}(r,c;\mathcal T)
=
1-
\frac{
\sum_{b\in\{0,1\}}
\frac{\eta(\mathcal T_b)}{\eta(\mathcal T)}
V(\mathcal T_b)
}{
V(\mathcal T)
},
\]
where each initialization is weighted inversely to the number of dates its winter contributes to $\mathcal T$, $\eta$ denotes the summed weight, and $V$ is the weighted trajectory variance averaged over NAM lead days 6--10.
For each candidate relationship $r$, we select the eligible threshold
\[
c_r^\star
=
\arg\max_c \mathrm{FVE}(r,c;\mathcal T).
\]
To retain reproducible relationships, we use 1{,}000 winter-level 70/30 train--held-out splits, refitting the threshold on training winters and requiring the separation direction to reproduce in at least 95\% of held-out splits~\citep{meinshausen_stability_2010}.
We then rerun the complete search under 100 cross-winter trajectory permutations and retain the smallest FVE threshold whose upper bootstrap bound (97.5th percentile) on the empirical false discovery rate (FDR) is at most 0.1\%~\citep{westfall_resampling-based_1993}.

This gives 283 first-level relationships and thresholds that are frozen, defining 566 branches.
An exhaustive second-level search within these branches applies the same stability and permutation-FDR criteria, yielding 672 retained conditional relationships. 
For each retained split, we select the terminal condition whose days-6–10 NAM mean deviates furthest from its parent branch mean and freeze its ERA5 initialization subset for model evaluation.
The same frozen first-level branches are used in every second-level null search. 
Full resampling and FDR definitions are given in Appendix~\ref{app:relationship-search}.

GraphCast, FengWu, and Pangu-Weather are evaluated on the same initialization subsets. 
For relationship $r$, model $m$, and lead time $\tau$, we measure
\[
E_{\mu}^{(r,m)}(\tau)
=
\left|
\mu_{m,r}(\tau)-\mu_{\mathrm{ERA5},r}(\tau)
\right|,
\qquad
E_{\sigma^2}^{(r,m)}(\tau)
=
\left|
\sigma^2_{m,r}(\tau)-\sigma^2_{\mathrm{ERA5},r}(\tau)
\right|.
\]
Here $\mu_{m,r}(\tau)$ and $\sigma^2_{m,r}(\tau)$ are the winter-weighted mean and variance of model $m$'s forecast $N_{50:100}(t_c+\tau)$.
Conventional $T$, $u$, and $v$ forecast errors are evaluated on the same subsets, providing a direct comparison between state-space forecast accuracy and preservation of the ERA5 conditional dynamics.

\begin{figure}[t]
    \centering
    \begin{subfigure}[t]{0.32\textwidth}
        \centering
        \includegraphics[width=\linewidth]{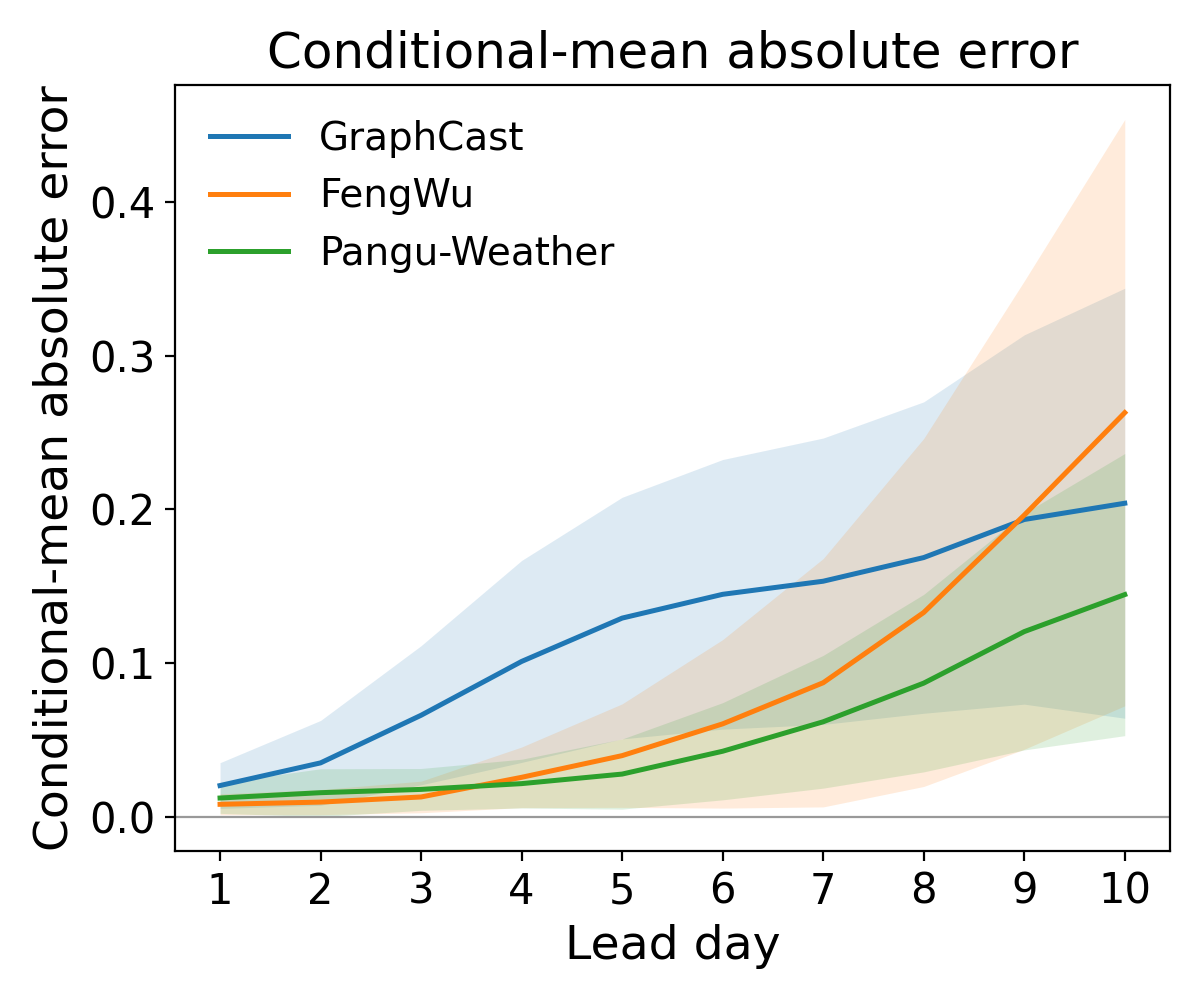}
        \caption{Conditional-mean error.}
        \label{fig:model-eval-mean}
    \end{subfigure}
    \hfill
    \begin{subfigure}[t]{0.32\textwidth}
        \centering
        \includegraphics[width=\linewidth]{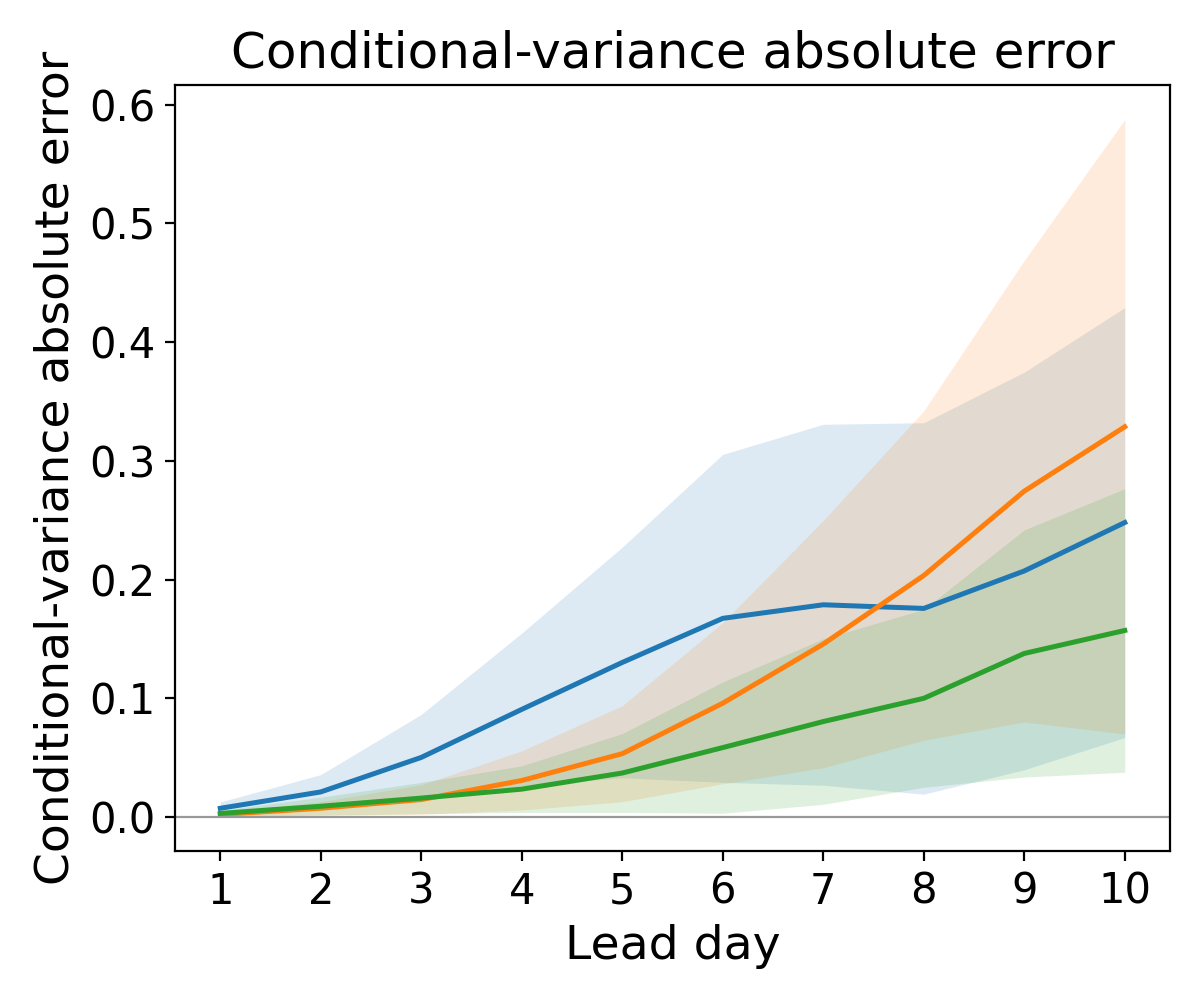}
        \caption{Conditional-variance error.}
        \label{fig:model-eval-variance}
    \end{subfigure}
    \hfill
    \begin{subfigure}[t]{0.32\textwidth}
        \centering
        \includegraphics[width=\linewidth]{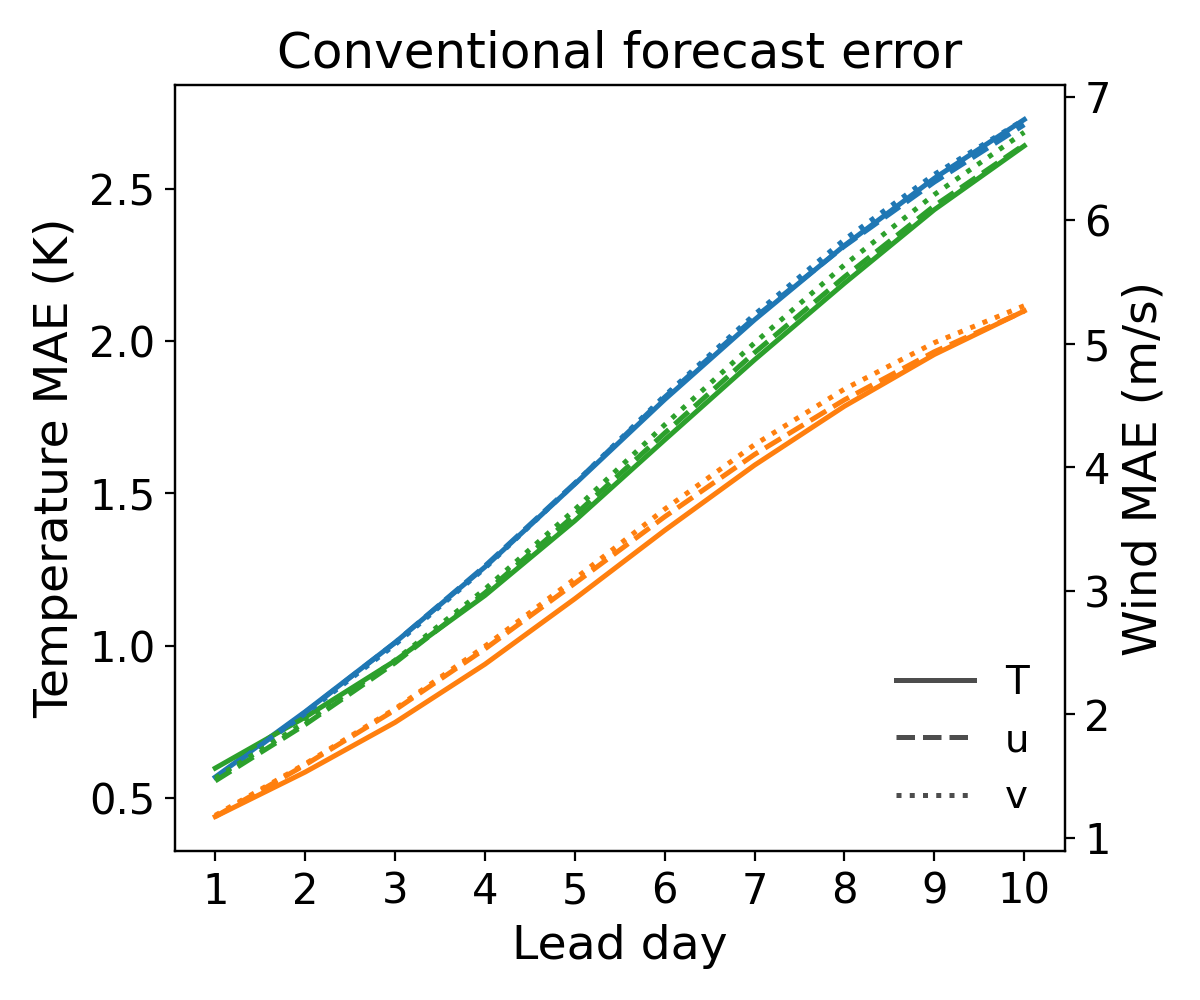}
        \caption{Conventional forecast error.}
        \label{fig:model-eval-conventional}
    \end{subfigure}

    \caption{%
        Model evaluation across the 672 selected terminal conditions associated with the retained second-level relationships.
        \textbf{(a--b)} Absolute error in the conditional mean and variance of future 50--100\,hPa NAM within each frozen ERA5-defined terminal leaf; lines show the mean across relationships and shading $\pm 1$ standard deviation across relationships.
        \textbf{(c)} Conventional $T$, $u$, and $v$ forecast MAE on the same subsets (unweighted global mean over the $1^\circ$ grid and 13 pressure levels).
    }
    \label{fig:model-evaluation}
\end{figure}

\section{ERA5 reveals sparse and conditional predictive structure}
\label{sec:results-landscape}
The ERA5 search reveals a highly structured predictive landscape for the subsequent evolution of initially neutral NAM states
(Figure~\ref{fig:predictive-landscape}).
Predictability is not distributed uniformly across the expert-defined representation space: it is concentrated in particular combinations of physical representation, spatial region, temporal window, and aggregation operator.
Representations of the background circulation and planetary-wave activity contain the strongest first-level signals, while the successive representation--wavenumber and temporal deep-dives show that these signals are localized to specific physical and temporal configurations rather than to broad diagnostic families as a whole.
This structure is also sparse when tested for reproducibility.
Only 295 of the 7{,}875 first-level candidates reproduce their separation direction across winters, with 283 surviving the permutation calibration (Figure~\ref{fig:selection}a).
More importantly, conditioning on these first-level states exposes substantial additional structure: exhaustive search within their frozen branches identifies 672 robust second-level relationships from more than 4.3 million conditional candidates (Figure~\ref{fig:selection}b).
Thus, much of the reference predictability is not captured by a single precursor in isolation, but emerges only after conditioning on another aspect of the preceding circulation.

The retained relationships therefore describe multiple physically distinct routes from comparable initial NAM states toward different subsequent circulation evolutions. 
Figure~\ref{fig:selection-best-leaf} illustrates one such conditional configuration leading to strongly negative subsequent NAM.
Taken together, the results depict the reference system not as a small set of prescribed process diagnostics, but as a sparse, hierarchical landscape of predictive relationships distributed across wave activity, wave forcing, background circulation, and prior circulation evolution.
\begin{table*}[t]
\centering
\scriptsize
\setlength{\tabcolsep}{1.8pt}
\renewcommand{\arraystretch}{1.10}

\caption{
Model evaluation over lead days 6--10 for selected subsets of the 672 selected terminal conditions.
Rows are defined by representations used in at least one split and may therefore overlap.
$E_\mu$ and $E_{\sigma^2}$ denote conditional NAM mean and variance errors; $T$, $u$, and $v$ are conventional forecast MAE (unweighted global mean over the $1^\circ$ grid and 13 pressure levels).
\best{Dark green} and \second{light green} indicate the lowest and second-lowest model errors in each column, respectively.
}
\label{tab:representation-model-evaluation}

\begin{tabular*}{\textwidth}{
    @{\extracolsep{\fill}}
    llrr
    ccccc
    ccccc
    ccccc
    @{}
}
\toprule
&&&&
\multicolumn{5}{c}{\textbf{GraphCast}}
& \multicolumn{5}{c}{\textbf{FengWu}}
& \multicolumn{5}{c}{\textbf{Pangu-Weather}} \\
\cmidrule(lr){5-9}
\cmidrule(lr){10-14}
\cmidrule(lr){15-19}

&&&&
\multicolumn{2}{c}{Fidelity}
& \multicolumn{3}{c}{Forecast}
& \multicolumn{2}{c}{Fidelity}
& \multicolumn{3}{c}{Forecast}
& \multicolumn{2}{c}{Fidelity}
& \multicolumn{3}{c}{Forecast} \\
\cmidrule(lr){5-6}
\cmidrule(lr){7-9}
\cmidrule(lr){10-11}
\cmidrule(lr){12-14}
\cmidrule(lr){15-16}
\cmidrule(lr){17-19}

&
\textbf{Predictive structure}
& $\mathbf{N}$
& \textbf{\%}
& $E_\mu$ & $E_{\sigma^2}$ & $T$ & $u$ & $v$
& $E_\mu$ & $E_{\sigma^2}$ & $T$ & $u$ & $v$
& $E_\mu$ & $E_{\sigma^2}$ & $T$ & $u$ & $v$ \\
\midrule

        & All
        & 672 & 100.0
        & 0.17 & \second{0.20} & 2.29 & 5.73 & 5.77
        & \second{0.15} & 0.21 & \best{1.76} & \best{4.49} & \best{4.56}
        & \best{0.09} & \best{0.11} & \second{2.17} & \second{5.49} & \second{5.58} \\

        \midrule

        \multirow{11}{*}{\textbf{Representations}}
        & Total EP1
        & 102 & 15.2
        & 0.21 & \second{0.18} & 2.30 & 5.74 & 5.76
        & \second{0.18} & 0.20 & \best{1.77} & \best{4.50} & \best{4.57}
        & \best{0.09} & \best{0.10} & \second{2.19} & \second{5.53} & \second{5.61} \\

        & Wave-1 EP1
        & 96 & 14.3
        & 0.15 & 0.26 & 2.29 & 5.71 & 5.77
        & \second{0.11} & \second{0.21} & \best{1.77} & \best{4.50} & \best{4.57}
        & \best{0.09} & \best{0.11} & \second{2.16} & \second{5.45} & \second{5.55} \\

        & Wave-2 EP1
        & 12 & 1.8
        & 0.26 & 0.20 & 2.27 & 5.68 & 5.69
        & \second{0.18} & \second{0.17} & \best{1.75} & \best{4.45} & \best{4.52}
        & \best{0.11} & \best{0.07} & \second{2.17} & \second{5.47} & \second{5.55} \\

        & Total EP2
        & 12 & 1.8
        & \second{0.18} & 0.20 & 2.30 & 5.75 & 5.78
        & 0.20 & \second{0.18} & \best{1.78} & \best{4.51} & \best{4.59}
        & \best{0.11} & \best{0.12} & \second{2.20} & \second{5.54} & \second{5.61} \\

        & Wave-1 EP2
        & 239 & 35.6
        & \second{0.17} & \second{0.19} & 2.30 & 5.75 & 5.79
        & 0.17 & 0.23 & \best{1.76} & \best{4.49} & \best{4.55}
        & \best{0.11} & \best{0.10} & \second{2.17} & \second{5.49} & \second{5.57} \\

        & Wave-2 EP2
        & 27 & 4.0
        & 0.21 & \best{0.10} & 2.29 & 5.71 & 5.73
        & \second{0.19} & 0.14 & \best{1.78} & \best{4.51} & \best{4.60}
        & \best{0.11} & \second{0.10} & \second{2.18} & \second{5.49} & \second{5.59} \\

        & Total Div
        & 48 & 7.1
        & \second{0.17} & \second{0.19} & 2.30 & 5.74 & 5.76
        & 0.20 & 0.27 & \best{1.78} & \best{4.51} & \best{4.57}
        & \best{0.12} & \best{0.09} & \second{2.20} & \second{5.52} & \second{5.60} \\

        & Wave-1 Div
        & 146 & 21.7
        & \second{0.20} & \second{0.15} & 2.30 & 5.74 & 5.77
        & 0.25 & 0.28 & \best{1.78} & \best{4.51} & \best{4.57}
        & \best{0.12} & \best{0.11} & \second{2.20} & \second{5.53} & \second{5.59} \\

        & Wave-2 Div
        & 57 & 8.5
        & 0.12 & 0.15 & 2.26 & 5.67 & 5.71
        & \second{0.09} & \second{0.14} & \best{1.76} & \best{4.48} & \best{4.55}
        & \best{0.07} & \best{0.10} & \second{2.17} & \second{5.48} & \second{5.59} \\

        & Zonal-mean $T$
        & 135 & 20.1
        & 0.17 & 0.20 & 2.27 & 5.69 & 5.73
        & \second{0.10} & \second{0.18} & \best{1.74} & \best{4.46} & \best{4.54}
        & \best{0.07} & \best{0.14} & \second{2.18} & \second{5.50} & \second{5.60} \\

        & Zonal-mean $u$
        & 410 & 61.0
        & 0.15 & 0.22 & 2.29 & 5.73 & 5.78
        & \second{0.10} & \second{0.19} & \best{1.75} & \best{4.48} & \best{4.55}
        & \best{0.08} & \best{0.11} & \second{2.16} & \second{5.48} & \second{5.57} \\

        \midrule

        \multirow{4}{*}{\textbf{Evolution}}
        & Weak positive shift
        & 61 & 9.1
        & 0.14 & 0.14 & 2.28 & 5.71 & 5.76
        & \second{0.09} & \second{0.11} & \best{1.75} & \best{4.46} & \best{4.53}
        & \best{0.08} & \best{0.09} & \second{2.15} & \second{5.47} & \second{5.56} \\

        & Weak negative shift
        & 107 & 15.9
        & 0.13 & 0.17 & 2.29 & 5.73 & 5.79
        & \second{0.09} & \second{0.16} & \best{1.75} & \best{4.47} & \best{4.54}
        & \best{0.08} & \best{0.09} & \second{2.14} & \second{5.45} & \second{5.54} \\

        & Strong positive shift
        & 19 & 2.8
        & 0.12 & 0.14 & 2.23 & 5.59 & 5.59
        & \second{0.06} & \second{0.08} & \best{1.72} & \best{4.40} & \best{4.48}
        & \best{0.05} & \best{0.07} & \second{2.12} & \second{5.39} & \second{5.47} \\

        & Strong negative shift
        & 150 & 22.3
        & \second{0.27} & \second{0.16} & 2.32 & 5.77 & 5.77
        & 0.33 & 0.34 & \best{1.79} & \best{4.53} & \best{4.59}
        & \best{0.15} & \best{0.12} & \second{2.21} & \second{5.53} & \second{5.58} \\

\bottomrule
\end{tabular*}
\end{table*}

\section{Dynamical fidelity in ML forecasts}
\label{sec:model-eval}

We next test whether GraphCast, FengWu, and Pangu-Weather preserve the ERA5-discovered conditional evolution when initialized from the same states and evaluated on the same frozen subsets.
Figure~\ref{fig:model-evaluation} shows that model performance differs depending on whether it is measured by state-space error or preservation of the reference predictive structure.
FengWu reproduces the conditional NAM mean and variance most closely over the first three lead days and achieves the lowest $T$, $u$, and $v$ MAE at every lead time, but its conditional errors grow fastest and exceed those of both other models by day 10. 
From day 4 onward, Pangu-Weather most closely reproduces the conditional NAM mean and variance.
Thus, lower state-space forecast error does not imply closer agreement with the conditional dynamics identified in ERA5.

More importantly, dynamical fidelity is not uniform across predictive relationships.
Table~\ref{tab:representation-model-evaluation} shows that Pangu-Weather has the lowest conditional-mean error across the reported representation and evolution subsets, while FengWu retains the lowest conventional forecast error.
The relative fidelity of GraphCast and FengWu, however, changes substantially with the predictive structure being evaluated.
For strong negative NAM shifts, GraphCast has lower conditional-mean error than FengWu ($0.27$ versus $0.33$), reversing their ordering for weak negative shifts ($0.13$ versus $0.09$), although FengWu has the lower NAM forecast error on both subsets (Appendix~\ref{app:ml_models}).
Similar variation occurs across individual wave-propagation and wave-forcing representations.
These results show that dynamical fidelity is not a single model property: models can preserve different parts of the reference predictive structure even when their aggregate forecast errors are similar. 
Conventional forecast skill can therefore conceal where within the physical dynamics model behaviour agrees with, or departs from, the reference system.

\section{Conclusion}
We introduced a framework for evaluating learned physical dynamics through predictive structure in expert-defined representation spaces. Experts define the physical vocabulary, while robust relationships are discovered from reference trajectories and frozen as evaluation tests.
Applied to stratospheric circulation, the framework reveals differences in how Pangu-Weather, GraphCast, and FengWu preserve ERA5 wave--mean-flow relationships and subsequent NAM evolution that are not captured by conventional forecast errors. 
Accurate state prediction therefore does not by itself establish preservation of reference predictive structure.
These relationships should not be interpreted as identified physical mechanisms: ERA5 serves as the reference rather than ground truth, and the framework evaluates predictive rather than causal structure. 
Because some precursors summarize histories longer than those directly available to the forecast models, disagreement may reflect either missing initialization information or incorrect subsequent evolution. 
Finally, stability filtering and permutation calibration reduce sensitivity to chance relationships within the finite ERA5 winter cohort but do not establish transfer to other periods, reanalyses, or physical systems.

\section*{Reproducibility Statement}
The relationship catalogue, discovery procedure, and dynamical-fidelity and conventional forecast diagnostics are specified in Appendices~\ref{app:wave_theory}-~\ref{app:ml_models}.
Anonymized code to reproduce the ERA5 inputs, relationship discovery, model forecasts, evaluation, and figures is available \href{https://anonymous.4open.science/r/discover-process-level-dynamical-fidelity-C5C0/}{here}.

\section*{AI use statement}
Generative AI tools, including OpenAI ChatGPT and Anthropic Claude, were used throughout this work to support methodological development, mathematical formulation, code development and review, interpretation of results, literature-oriented brainstorming and search, and drafting and editing of the manuscript.
AI-assisted code and analyses were tested and verified by the authors, and literature and scientific claims were checked against the underlying sources.
The authors made all final decisions regarding the methodology, experiments, interpretation, and presentation of results, and take responsibility for the full content of the work.

\bibliographystyle{unsrtnat}
\bibliography{references}  

\newpage

\appendix

\section{Budget and algebraic constraints as coarsenings of the differential residual}
\label{app:coarsening}
This appendix makes the evaluation axes of Section~\ref{sec:dynamical-fidelity} precise and shows how they relate. 
We express every test through a single quantity, the differential residual, which measures how far a trajectory departs from the reference dynamics. 
This lets us order the tests by what they can detect,~\eqref{eq:budget}, and a worked example then shows that physical consistency and process-level dynamical fidelity capture different aspects of the dynamics and are therefore complementary.

Let the reference dynamics be \(\dot{x}=F(x)\). The learned model is instead a discrete map \(f_\theta\) that advances the state $x(t)$ by one step \(\Delta t\).
A model trained on states does not represent $F$ directly, but its integral over $\Delta t$.

The issue of not directly modeling $F$ concerns the discrete-time form of the model, not the resolution of the state, and for numerical one-step methods with a small step, backward error analysis shows that the agreement between the discrete and continuous model can be made up to an error exponentially small in \(1/\Delta t\)~\citep{hairer_geometric_2006}.

The purpose of the discrete-time dynamics assumption is to give the predicted trajectory a meaning between grid times, so that reference and prediction can be tested with the same quantity. 
We then define $Q_0$, one of the $Q_\alpha$, abusing notation here to include trajectories in its domain when it took the vector field originally, due to the discretization relation discussed above, as the differential residual
\begin{equation}
    Q_0[x](t)=\dot{x}(t)-F\bigl(x(t)\bigr),
\end{equation}
which compares the actual rate of change of a trajectory \(x\) with the rate that the reference dynamics prescribes at the same state. 
For a trajectory of the reference system the two coincide by definition, so \(Q_0[x]\equiv0\), and a non-zero residual signals a trajectory that the reference system could not produce. 
Along a predicted trajectory \(\dot{\widehat{x}}=\widehat{F}(\widehat{x})\), so the residual equals the error of the learned vector field, \(\delta F=\widehat{F}-F\), at the visited states,
\begin{equation}
    Q_0[\widehat{x}](t)=\delta F\bigl(\widehat{x}(t)\bigr).
    \label{eq:diffres}
\end{equation}
Tests evaluated on predicted trajectories therefore see \(\delta F\) only at the states the model visits, much as a test set probes a model only on the inputs it contains.
 
In practice, the relations \(Q_j\) of Section~\ref{sec:dynamical-fidelity} are checked on trajectories, often sampled coarsely in space and time. 
We could write \(Q_j[x]\) for relation \(j\) evaluated on a trajectory \(x\), with \(Q_j[x]=0\) whenever \(x\) is a reference trajectory. 
The residual \(Q_0\) is the vector field relation in trajectories form, and the relations below are all expressed through it.
A set of relations \(J\) is weaker than \(J'\), written \(J\preceq J'\), if every trajectory satisfying the relations in \(J'\) also satisfies those in \(J\), as \(\mathfrak{F}_{J'}\subseteq\mathfrak{F}_J\) in Section~\ref{sec:dynamical-fidelity}, and strictly weaker, \(J\prec J'\), if some trajectory satisfies \(J\) but not \(J'\).
Since a trajectory with \(Q_0[x]\equiv0\) is a reference trajectory, \(Q_0\) is stronger than every other relation.
\begin{assumption}[Flow representation]
\label{ass:flow}
The learned map is the one-step flow of a vector field \(\widehat F\in\mathfrak F\), that is, \(f_\theta(\widehat x_t)=\widehat x_{t+\Delta t}\) with \(\dot{\widehat x}=\widehat F(\widehat x)\) in between.
\end{assumption}

\begin{proposition}[Ordering of evaluation tests]
\label{prop:ladder}
Under Assumption~\ref{ass:flow}, let \(F\) be Lipschitz, and consider finitely many invariants \(\{C_j\}\), finitely many budgets \(\{B_j\}\) over consecutive intervals of fixed length \(\Delta t\) that include the budgets of the invariants, and finitely many process diagnostics \(\{\mathcal{E}^{\mathrm{dyn}}_k\}\) evaluated from the reference initial state. Then
\begin{equation}
\label{eq:budget}
\underbrace{\{C_j\}}_{\text{invariants}}
\;\preceq\;
\underbrace{\{B_j\}}_{\text{budgets}}
\;\preceq\;
\underbrace{Q_0}_{\text{differential residual}}
\;\succeq\;
\underbrace{\{\mathcal{E}^{\mathrm{dyn}}_k\}}_{\text{process diagnostics}},
\end{equation}
where each relation can be strict under the following conditions:
\begin{enumerate}
    \item Invariants to budget, is strict whenever the budgets include a quantity that the dynamics changes.
    \item Process diagnostics and invariants are not ordered: a trajectory can pass every invariant and fail a process diagnostic, or pass a process diagnostic and fail an invariant, and therefore, since an invariant is a sum of budgets, at least one budget of that quantity.
\end{enumerate}

\end{proposition}
\noindent\emph{Proof.} Sections~\ref{subsec:budgets} and~\ref{subsec:invariants} establish the chain on the left and Section~\ref{subsec:process} the relation on the right. For strictness and non-comparability, the scaled oscillator below passes every invariant but fails a budget and the period diagnostic, the damped oscillator passes the period diagnostic but fails the energy invariant, and a residual whose weighted integrals vanish on every interval passes all budgets while \(Q_0\neq0\) (Section~\ref{subsec:budgets}).

\subsection{Budgets} 
\label{subsec:budgets}
A budget states that the change of some quantity over a time interval equals what the dynamics adds or removes during that interval, as when the change in heat content of an ocean region equals the net heat flux through its boundaries~\citep{peixoto_physics_1992}.
Our aim here is to write this statement as a relation in the sense of Section~\ref{sec:dynamical-fidelity}, a residual that vanishes on reference trajectories, and to express it through \(Q_0\). This form shows exactly what a budget checks, and what it cannot see.
The budgeted quantity is a functional \(E_j\), a number computed from the whole state, such as the mass \(E_j(u)=\int_{\Omega'}u\,\mathrm{d}z\) contained in a region \(\Omega'\) of the domain \(\Omega\), or the kinetic energy \(E_j(u)=\tfrac12\int_\Omega|u|^2\,\mathrm{d}z\). \(E_j\) is the quantity being tracked, not a relation; the relation is the budget \(B_j\) defined below, which plays the role of \(Q_j[x]\).
 
To describe how \(E_j\) changes, we use the spatial inner product \(\langle\varphi,v\rangle=\int_\Omega\varphi\cdot v\,\mathrm{d}z\), which returns the \(\varphi\)-weighted total of a field \(v\) over the domain and reduces to the Euclidean dot product for finite-dimensional states. The gradient \(\partial_x E_j(x)\) is defined by \(E_j(x+\varepsilon v)=E_j(x)+\varepsilon\langle\partial_x E_j(x),v\rangle+o(\varepsilon)\), so that \(\langle\partial_x E_j(x),v\rangle\) is the change of \(E_j\) caused by a small change \(v\) of the state; it is \(\mathbf{1}_{\Omega'}\) for the regional mass and \(u\) for the kinetic energy. Along any trajectory, the chain rule gives \(\tfrac{\mathrm{d}}{\mathrm{d}t}E_j(x)=\langle\partial_x E_j(x),\dot{x}\rangle\), and integrating the residual weighted by this gradient over \([t,t+\Delta t]\) gives
\begin{equation}
\begin{aligned}
B_j&:=\int_t^{t+\Delta t}\!\bigl\langle\partial_x E_j(x(t')),Q_0[x](t')\bigr\rangle\,\mathrm{d}t'\\
&\phantom{:}=E_j(x_{t+\Delta t})-E_j(x_t)-\int_t^{t+\Delta t}\!\bigl\langle\partial_x E_j(x(t')),F(x(t'))\bigr\rangle\,\mathrm{d}t'.
\end{aligned}
\label{eq:budget_residual}
\end{equation}
The second line is the budget as it is computed in practice: the change in storage minus the tendency accumulated under the reference dynamics. For conservation-form dynamics \(F(u)=-\nabla\!\cdot f(u)\) and the regional mass, the tendency is the net inflow \(-\oint_{\partial\Omega'}f\cdot n\,\mathrm{d}S\), which recovers the familiar control-volume budget, storage change equals net inflow. Evaluating it requires \(F\) along the whole path within the interval, so from grid states alone it is computed by quadrature and vanishes only up to quadrature error. The first line is what places budgets relative to \(Q_0\): a budget is the residual averaged against the weight \(\partial_x E_j\) over the interval.
 
Budgets are therefore necessary conditions for the reference dynamics, but not sufficient ones.
They are necessary because \(B_j\) is built from \(Q_0\): if a trajectory follows the reference dynamics, then \(Q_0[x]\equiv0\) and every budget holds, so \(\{B_j\}\preceq Q_0\).
They are not sufficient because a budget sees the residual only after it has been summed, over the time interval and over the part of the state that \(E_j\) measures.
Errors that cancel in this sum go unnoticed. A transport that is too fast in the first half of a step and too slow in the second leaves the budgets over that step satisfied, and mass that moves to the wrong place inside \(\Omega'\) leaves the mass budget of \(\Omega'\) satisfied, although in both cases \(Q_0\neq0\).
Only checking every possible quantity \(E_j\) over arbitrarily short intervals would recover \(Q_0\), since a function whose weighted integrals over every interval all vanish is itself zero.
With finitely many budgets at a fixed \(\Delta t\), as in any practical evaluation, \(\{B_j\}\prec Q_0\).
 
\subsection{Invariants}
\label{subsec:invariants}
An invariant is a quantity \(I_j\) that the reference dynamics cannot change, such as the total mass, energy or momentum of a closed, unforced system. In terms of the gradient introduced for budgets, its tendency vanishes at every admissible state, \(\langle\partial_x I_j(x),F(x)\rangle=0\): the dynamics moves the state only in directions that leave \(I_j\) unchanged. An invariant is therefore a budget whose tendency term is identically zero. Taking \(E_j=I_j\) in Eq.~\eqref{eq:budget_residual} over \([t_0,t]\), the budget reduces to a comparison of two snapshots,
\begin{equation}
C_j(x_t):=I_j(x_t)-I_j(x_{t_0})=\int_{t_0}^{t}\bigl\langle\partial_x I_j(x(s)),Q_0[x](s)\bigr\rangle\,\mathrm{d}s .
\label{eq:invariant_residual}
\end{equation}
This is why invariants are cheap to check: they need only the states at grid times, not the reference tendency \(F\). The global mass budget of a closed domain is the invariant \(I_j(u)=\int_\Omega u\,\mathrm{d}z\). Instantaneous constraints such as \(\nabla\!\cdot u=0\) fit the same form: in incompressible dynamics the divergence is itself conserved, so it stays zero if it is zero initially.

Invariants are weaker than budgets. Since an invariant is a budget, a trajectory that satisfies the budgets of \(I_j\) over consecutive intervals covering \([t_0,t]\) also satisfies \(C_j\), their sum, so \(\{C_j\}\preceq\{B_j\}\). The converse fails because an invariant sees only the part of the residual that changes \(I_j\). A model that follows the reference dynamics at the wrong speed, \(\dot{x}=\lambda F(x)\) with \(\lambda\neq1\), has residual \(Q_0=(\lambda-1)F\), which changes no invariant, so every \(C_j\) stays zero. A budget of a quantity that the dynamics does change still registers the error, since along this trajectory
\begin{equation*}
B_j=(\lambda-1)\int_t^{t+\Delta t}\!\bigl\langle\partial_x E_j,F\bigr\rangle\,\mathrm{d}t'=\frac{\lambda-1}{\lambda}\bigl[E_j(x_{t+\Delta t})-E_j(x_t)\bigr],
\end{equation*}
which is non-zero whenever \(E_j\) changes over the interval. Hence \(\{C_j\}\prec\{B_j\}\) as soon as the budgets include such a quantity; the harmonic oscillator below makes this concrete.

\subsection{Process diagnostics}
\label{subsec:process}
A process diagnostic \(\mathcal{P}_k\) extracts one aspect of how the system evolves over a time window of length \(L\), such as the propagation speed of a wave, the rate at which a tracer is carried across a region, or the response to a change in forcing. Such diagnostics are standard in the evaluation of climate and weather models~\citep{maloney_process-oriented_2019} and have been applied to machine-learning weather models~\citep{hakim_dynamical_2024}.
The corresponding relation compares this aspect with that of the reference trajectory from the same initial state,
\begin{equation*}
    \mathcal{E}^{\mathrm{dyn}}_k=d_k\bigl(\mathcal{P}_k(\widehat{x}_{t:t+L}),\mathcal{P}_k(x_{t:t+L})\bigr),
\end{equation*}
which is zero whenever the two trajectories agree in the aspect that \(\mathcal{P}_k\) measures, and in particular whenever they coincide.

Process diagnostics are implied by \(Q_0\). If \(Q_0[\widehat{x}]\equiv0\) and \(\widehat{x}_{t_0}=x_{t_0}\), the prediction follows the reference dynamics from the reference initial state. When this determines a unique trajectory, as it does for Lipschitz \(F\), whose tendency changes at most proportionally to a change of state, then \(\widehat{x}=x\) and every \(\mathcal{E}^{\mathrm{dyn}}_k\) vanishes, so \(\{\mathcal{E}^{\mathrm{dyn}}_k\}\preceq Q_0\).
The converse fails because each diagnostic keeps only one aspect of the trajectory and discards the rest, so a model can match every diagnostic in the family while \(Q_0\neq0\). A diagnostic of oscillation period, for instance, is blind to errors in amplitude, as the damped oscillator of the example below shows. Hence \(\{\mathcal{E}^{\mathrm{dyn}}_k\}\prec Q_0\).

Process diagnostics also differ in kind from budgets and invariants. Budgets and invariants check that weighted sums of the residual \(Q_0\) are zero, whereas a process diagnostic checks an outcome of the dynamics, such as when a wave arrives. A residual can change that outcome while every weighted sum stays zero, and a residual that breaks a budget can leave the outcome unchanged. Neither kind of relation therefore implies the other, and the example below shows both cases.

\subsection{Example: harmonic oscillator} Consider the oscillator with state \(x=(q,p)\), \(F=(p,-q)\) and energy \(H=\tfrac12(q^2+p^2)\). From \(x_{t_0}=(1,0)\) the reference trajectory is \(x(t)=(\cos\tau,-\sin\tau)\), \(\tau=t-t_0\), a circle traversed with period \(2\pi\), and we take this period as the process diagnostic.

Suppose a surrogate model \(\widehat{F}_{\mathrm{scaled}}=\lambda F\), \(\lambda\neq1\), traverses the same circle at a different speed, \(\widehat{x}(t)=(\cos\lambda\tau,-\sin\lambda\tau)\). It preserves energy, phase-space volume and rotational symmetry exactly, and so passes every invariant, yet its period is \(2\pi/\lambda\), and the state error \(\|\widehat{x}(t)-x(t)\|=2|\sin((\lambda-1)\tau/2)|\) reaches its maximum of \(2\) at \(\tau=\pi/|\lambda-1|\), when the prediction lies on the opposite side of the circle. 
The budget of \(E_j(x)=q\) detects this error, \(B_j=(\lambda-1)\int_t^{t+\Delta t}\widehat{p}\,\mathrm{d}t'\neq0\) for generic slabs.

Consider now  a second surrogate, the damped model $\widehat{F}_{\mathrm{damped}}(x) = F(x)-\gamma x$,$\gamma>0$, which gives $\widehat{x}(t) = e^{-\gamma\tau} (\cos\tau,-\sin\tau)$. 
It oscillates with exactly the reference period and passes the diagnostic, but its energy decays as \(H(\widehat{x}(t))=\tfrac12e^{-2\gamma\tau}\), so it fails the energy invariant and every budget built on it.
The first model passes the invariants and fails the diagnostic, the second does the opposite, and neither satisfies \(Q_0\equiv0\).
The single invariant \(H\) thus fixes the circle on which the trajectory lies but not the speed at which it is traversed. 

More generally, \(m\) functionally independent invariants restrict \(\delta F\) only to the tangent space of their joint level set, and even \(n-1\) of them leave \(\delta F\) free along \(F\), so invariants restrict where a trajectory lies but not how it moves along it.
The residual penalised by a physics-informed loss~\citep{raissi_physics-informed_2019, karniadakis_physics-informed_2021} is in this sense the maximal element of the ladder, although it certifies \(\widehat{F}=F\) only on the states where it is evaluated.

\subsection{Remark on vector fields and flow maps.}
The analysis above assumes that the learned map is the one-step flow of some vector field, \(f_\theta(\widehat{x}_t)=\widehat{x}_{t+\Delta t}\) with \(\dot{\widehat{x}}=\widehat{F}(\widehat{x})\) in between. This is an idealisation: not every map arises in this way, and among diffeomorphisms of a compact manifold those that do are, in a precise sense, few~\citep{palis_vector_1974}.
A simple counterexample is the map \(x\mapsto -x\) on the real line. Trajectories of a flow in one dimension cannot cross, so its one-step map preserves the order of points, whereas \(x\mapsto -x\) reverses it.
The assumption is mild when the map is close to the identity. For numerical one-step methods with a small step, backward error analysis shows that the method is the exact flow of a modified vector field up to an error exponentially small in \(1/\Delta t\)~\citep{hairer_geometric_2006}. Machine-learning emulators, however, often use large steps, for which no such guarantee is available.
When the assumption fails, the relations above can still be evaluated with the finite difference \((\widehat{x}_{t+\Delta t}-\widehat{x}_t)/\Delta t\) in place of \(\dot{\widehat{x}}\), at the cost of the discretisation error discussed in Section~\ref{sec:dynamical-fidelity}.

\section{Hybrid and Physics-Informed Architectures}
\label{app:hybrid}

The tests of equation~\eqref{eq:budget} are evaluated on realised trajectories, whereas architectural and training-time constraints act on the learned vector field \(\widehat{F}\) or on the loss. Since a test sees the model only through \(\delta F\) at visited states, equation~\eqref{eq:diffres}, a constraint guarantees it by construction only if it forces the corresponding weighted error to vanish at every state, as \(\langle\partial_x I_j(x),\delta F(x)\rangle=0\) does for an invariant.

\subsection{Hybrid models} A differentiable-solver model~\citep{kochkov_neural_2024} sets \(\widehat{F}=F_{\mathrm{res}}+N_\theta\), where \(F_{\mathrm{res}}\) discretises the known equations on the resolved scales and \(N_\theta\) is a learned closure for the unresolved ones. Splitting the reference as \(F=F_{\mathrm{res}}+F_{\mathrm{sub}}\), with \(F_{\mathrm{sub}}\) the unresolved tendency plus the discretisation error, the model error is the closure error, \(\delta F=N_\theta-F_{\mathrm{sub}}\).
If \(F_{\mathrm{res}}\) conserves an invariant \(I_j\), so does \(F_{\mathrm{sub}}\), and \(C_j=\int_{t_0}^{t}\langle\partial_x I_j,N_\theta\rangle\,\mathrm{d}s\) vanishes by construction whenever the closure cannot change \(I_j\), as for total mass under a closure in flux form. A budget, \(B_j=\int_t^{t+\Delta t}\langle\partial_x E_j,N_\theta-F_{\mathrm{sub}}\rangle\,\mathrm{d}t'\), vanishes only if the closure reproduces the unresolved tendency seen by \(E_j\), such as the subgrid flux through the boundary of a region, which is learned rather than structural. This is the sense in which Section~\ref{sec:dynamical-fidelity} cites hybrid models as physically consistent.
By the argument of Appendix~\ref{app:coarsening}, these invariants constrain \(\delta F\) only along \(\partial_x I_j\), leaving the rest of the closure error, and with it the trajectory, free. Process diagnostics typically target the coupling between resolved dynamics and closure, as in wave--mean-flow interaction and the circulation response it drives, and so probe precisely the component no solver fixes. A hybrid model is thus expected to pass the constraints its structure enforces while remaining open at the process level, and the framework makes that expectation testable rather than assumed.

\subsection{Constrained architectures} Hard-constraint layers correct each predicted state so that chosen invariants or budgets hold, and pass those \(C_j\) or \(B_j\) exactly at grid times~\citep{hansen_learning_2023}, leaving every untested direction of \(\delta F\) free.
Equivariant architectures~\citep{batzner_e3-equivariant_2022} impose \(\widehat{F}(gx)=g\widehat{F}(x)\) for a symmetry group, which excludes errors that break the symmetry but not those that respect it: in the oscillator of Appendix~\ref{app:coarsening}, \(\widehat{F}=\lambda F\) is rotation-equivariant for every \(\lambda\).
Hamiltonian networks~\citep{greydanus_hamiltonian_2019} write \(\widehat{F}=J\nabla\widehat{H}\), with \(J\) the canonical symplectic matrix, and conserve the learned energy \(\widehat{H}\) exactly, but the reference energy only if \(\nabla H^\top J\nabla\widehat{H}=0\). Even then the trajectory is not fixed, since \(\lambda F=J\nabla(\lambda H)\) is Hamiltonian and conserves \(H\).

\subsection{Physics-informed losses} These losses penalise the residual \(Q_0\) itself, the strongest test of equation~\eqref{eq:budget}, but only where they are evaluated~\citep{karniadakis_physics-informed_2021}. At a training state the model residual is \(\delta F(x)\), so a small loss bounds \(\|\delta F\|^2\) on average over \(p_{\mathrm{train}}\), not on the states of a rollout, which drift away from \(p_{\mathrm{train}}\) as errors accumulate. Computed from grid states, the residual is moreover itself a coarsening of \(Q_0\).

\section{Planetary-Wave Diagnostics for Dynamical Fidelity}
\label{app:wave_theory}
This appendix elaborates on the definitions of the zonal-mean and eddy quantities, then the Eliassen--Palm (EP) fluxes and lastly the Northern Annular Mode (NAM).
Each diagnostic is well defined independently, however, during a sudden stratospheric warming (SSW), their expected temporal ordering links anomalous wave activity, disruption of the polar vortex, and the subsequent evolution of the circulation through the atmospheric column.

\label{app:zonal-decomposition}
\begin{figure}[h]
    \centering
    \includegraphics[width=\textwidth]{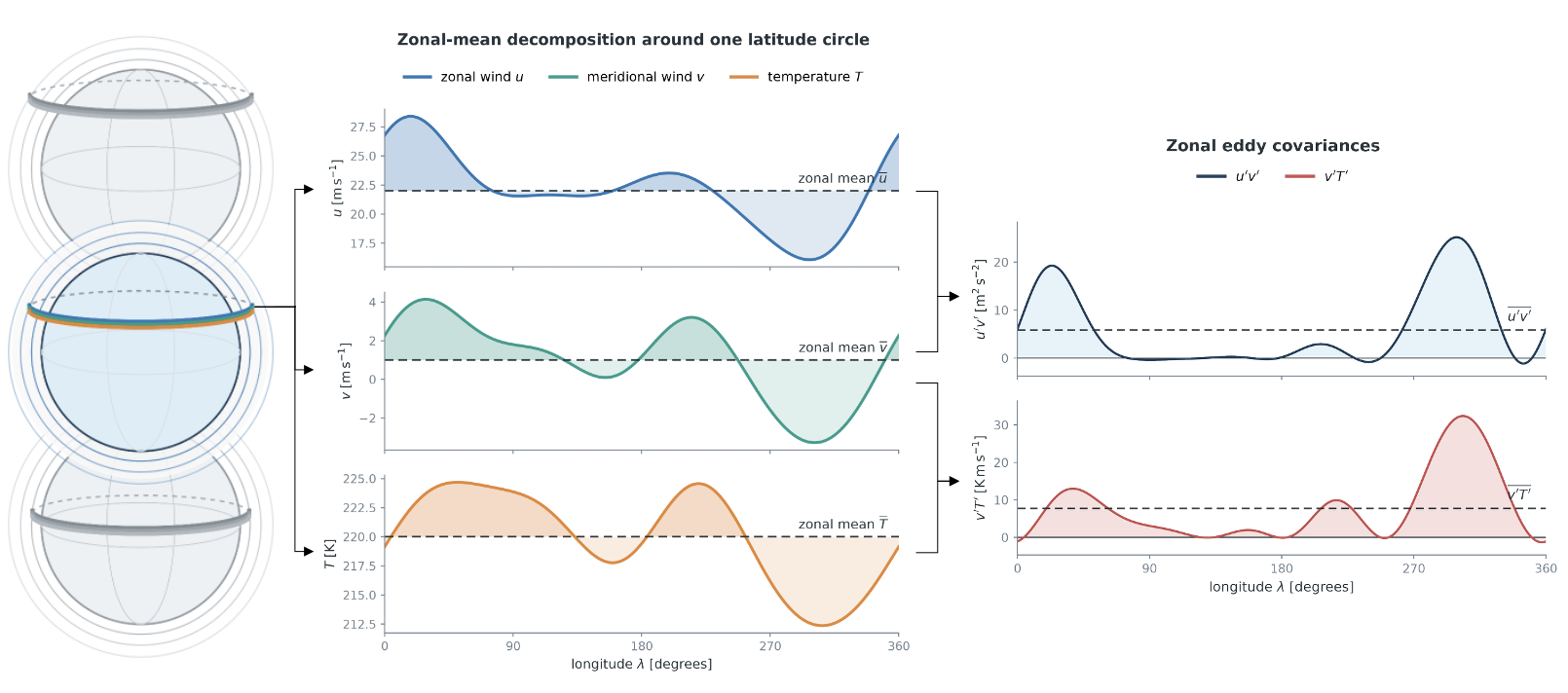}
    \caption{
    Schematic illustration of the zonal-mean decomposition and zonal eddy covariances. At a fixed latitude, pressure, and time, each atmospheric variable forms a longitude-dependent field around an east--west circle. Dashed lines show the zonal means, while the shaded departures are the eddies \(u'\), \(v'\), and \(T'\). Multiplying the relevant departures longitude by longitude and averaging the products around the circle gives \(\overline{u'v'}\) and \(\overline{v'T'}\). The faded globes indicate that the same calculation is repeated independently across latitude and pressure. The calculation is also repeated for every time-step but this is not included in the illustration above.
    }
    \label{fig:zonal-mean-explainer}
\end{figure}

\subsection{Zonal Means and Eddy Covariances}
The diagnostics are constructed from zonal wind \(u\), meridional wind \(v\), temperature \(T\), and geopotential height \(Z\), defined on a longitude--latitude--pressure grid. 
Longitude, latitude, pressure, and time are denoted by \(\lambda\), \(\phi\), \(p\), and \(t\), respectively.
Pressure is used as the vertical coordinate, so decreasing \(p\) corresponds to increasing altitude.

Planetary waves appear as longitude-dependent disturbances on a longitude-independent background circulation. 
We therefore average over longitude while retaining latitude, pressure, and time. 
This decomposition gives the wave pattern around every latitude circle and atmospheric level at every time. 
Examining the wave-change across pressure can reveal whether a pattern remains confined to the troposphere or extends into the stratosphere whereas examining the wave-change across time can reveal how the wave propagates through time.

For \(q\in\{u,v,T\}\), the discretized zonal mean with \(N_\lambda\) equally spaced longitude points becomes
\[
\overline q(\phi_j,p_k,t_n)
=
\frac{1}{N_\lambda}
\sum_{i=1}^{N_\lambda}
q(\lambda_i,\phi_j,p_k,t_n).
\]

The remaining longitude-dependent component is
\[
q'(\lambda,\phi,p,t)
=
q(\lambda,\phi,p,t)-\overline q(\phi,p,t),
\]
where \(q'\) collects all longitude-dependent departures from the zonal mean, from planetary-scale disturbances to smaller synoptic structures. Eddy refers specifically to this zonal-mean departure; no other mean--eddy decompositions are considered.

The dynamical importance of these eddies lies not only in their individual amplitudes or spatial scales, but in the transport of heat and momentum produced by their joint organization across variables. 
These transports redistribute heat between latitudes, help maintain and shift the jet streams and storm tracks, and mediate the wave forcing of the stratospheric polar vortex, making their faithful representation consequential for circulation changes such as SSWs. 
Crucially, no single-variable eddy field determines a transport: one variable specifies the anomalous motion, while another specifies the quantity carried by that motion.

We consider the two zonal eddy covariances
\[
\overline{v'T'}(\phi,p,t)
=
\frac{1}{N_\lambda}
\sum_{i=1}^{N_\lambda}
v'(\lambda_i,\phi,p,t)T'(\lambda_i,\phi,p,t),
\]
and
\[
\overline{u'v'}(\phi,p,t)
=
\frac{1}{N_\lambda}
\sum_{i=1}^{N_\lambda}
u'(\lambda_i,\phi,p,t)v'(\lambda_i,\phi,p,t).
\]
These quantities describe the meridional eddy transport of heat and zonal momentum, respectively. 
Their values depend jointly on the amplitudes and relative longitudinal positions of the disturbances in each variable. 
Consequently, a model may produce plausible wave structure in \(u\), \(v\), and \(T\) separately while suppressing or reversing the transport arising from their interaction.

\subsection{Fourier Interpretation}
\label{app:fourier-eddies}

Figure~\ref{fig:zonal-mean-explainer} illustrates the zonal-mean decomposition and eddy covariances directly in longitude space. The same operations can be interpreted in wavenumber space using a Fourier transform. A Fourier transform does not alter the underlying field; it represents the longitude-dependent pattern as a sum of waves with different spatial scales, amplitudes, and longitudinal positions.

At fixed latitude \(\phi\), pressure \(p\), and time \(t\), the eddy field \(q'(\lambda,\phi,p,t)\) is periodic in longitude and has zero zonal mean. It can therefore be written as
\[
q'(\lambda,\phi,p,t)
=
\sum_{m=1}^{M}
A_{q,m}(\phi,p,t)
\cos\!\left[
m\lambda-\alpha_{q,m}(\phi,p,t)
\right],
\]
where longitude \(\lambda\) is expressed in radians. The zonal wavenumber \(m\) counts the number of complete wave cycles around a latitude circle: \(m=1\) represents one cycle around the globe, \(m=2\) represents two, and progressively larger values represent smaller longitudinal scales. The \(m=0\) component is absent because it is precisely the zonal mean removed in the definition of \(q'\). The upper limit \(M\) is the largest wavenumber represented by the discrete longitude grid.

The coefficient \(A_{q,m}(\phi,p,t)\) is the amplitude of wavenumber \(m\) and has the same physical units as \(q\). The phase \(\alpha_{q,m}(\phi,p,t)\) determines the longitude of its crests and troughs. For the convention above, a crest occurs at
\[
\lambda_{\max}
=
\frac{\alpha_{q,m}}{m}
\quad
\left(\operatorname{mod}\frac{2\pi}{m}\right).
\]
Both amplitude and phase may vary with latitude, pressure, and time, allowing the strength and position of a given wave component to change throughout the atmospheric column and forecast trajectory.

The angular wavelength of wavenumber \(m\) is \(2\pi/m\), while its physical wavelength along a latitude circle is
\[
L_m(\phi)
=
\frac{2\pi a\cos\phi}{m},
\]
where \(a\) is Earth's radius. Low zonal wavenumbers therefore correspond to planetary-scale structures, while higher wavenumbers correspond to progressively smaller disturbances. Low-wavenumber waves often provide the dominant contribution to stratospheric wave propagation and polar-vortex forcing, whereas higher-wavenumber disturbances are commonly associated with tropospheric weather systems, fronts, and storm tracks.

The Fourier representation also makes the role of cross-variable alignment in the eddy covariances explicit. Consider two eddy fields \(a'\) and \(b'\) containing the same zonal wavenumber \(m\):
\[
a'_m
=
A_{a,m}\cos(m\lambda-\alpha_{a,m}),
\qquad
b'_m
=
A_{b,m}\cos(m\lambda-\alpha_{b,m}).
\]
Their zonally averaged product is
\[
\overline{a'_m b'_m}
=
\frac{1}{2}
A_{a,m}A_{b,m}
\cos\!\left(
\alpha_{a,m}-\alpha_{b,m}
\right).
\]
The resulting covariance is controlled jointly by the amplitudes of the two waves and their relative phase. Equal phases give the largest positive covariance, a phase difference of \(\pi/2\) causes positive and negative products to cancel around the latitude circle, and a phase difference of \(\pi\) reverses the sign of the covariance. Two individually strong waves can therefore produce little or no net transport when their longitudinal structures are incorrectly aligned.

Fourier modes with different wavenumbers are orthogonal around a complete latitude circle, so their cross-products vanish under the zonal mean. For eddy fields containing several wavenumbers, the total covariance is therefore the sum of the matching-wavenumber contributions, 
\[
\overline{a'b'}
=
\frac{1}{2}
\sum_{m=1}^{M}
A_{a,m}A_{b,m}
\cos\!\left(
\alpha_{a,m}-\alpha_{b,m}
\right).
\]
This decomposition can be applied to \(\overline{v'T'}\) or \(\overline{u'v'}\) to determine which zonal scales contribute to the meridional transport of heat or momentum. It can therefore distinguish an error in wave amplitude from an error in cross-variable phase, and identify whether a forecast failure originates primarily from
planetary-scale or smaller-scale disturbances. Unless stated otherwise, the diagnostics in the main text use the total covariance summed across all resolved zonal wavenumbers.

\subsection{Eliassen--Palm Flux and Wave Forcing}
\label{app:wave-diagnostics}

We combine the two eddy covariances into the quasi-geostrophic Eliassen--Palm (EP) flux \citep{edmon_eliassen-palm_1980},
\[
\mathbf{F}(\phi,p,t)
=
\left(F_\phi,F_p\right),
\qquad
F_\phi\propto-\overline{u'v'},
\qquad
F_p\propto\overline{v'T'}.
\]
The connection between heat transport and vertical propagation follows from the geometry of pressure surfaces: warm layers increase the distance between neighbouring pressure levels, whereas cold layers decrease it. The alignment of these thickness anomalies with north--south wind anomalies therefore reveals how the wave pattern tilts between pressure levels, which determines its vertical direction of propagation.

The EP flux produces a vector at every latitude, pressure level, and time.
Under the quasi-geostrophic Rossby-wave interpretation, \(F_\phi\) describes meridional propagation, while \(F_p\) describes propagation in pressure. 
Because pressure decreases with altitude, flux directed toward lower pressure corresponds to upward propagation. 

The EP-flux vector shows where wave activity is directed, but not whether it changes the circulation through which it travels. 
To measure this interaction, we compare how much wave activity enters and  leaves each small region of latitude--pressure space using the EP-flux divergence. 
Taking $F_p$ positive upward, i.e.\ toward lower pressure,
\[
\mathbf F
=
a\cos\phi\left(
-\overline{u'v'},\;
-\frac{f\,\overline{v'\theta'}}{\partial\overline\theta/\partial p}
\right),
\qquad
\nabla\!\cdot\!\mathbf{F}
=
\frac{1}{a\cos\phi}
\frac{\partial\!\left(F_\phi\cos\phi\right)}{\partial\phi}
-
\frac{\partial F_p}{\partial p},
\]
where $a$ is Earth's radius, $f$ the Coriolis parameter, and $\theta$ potential temperature. 
Because $\partial\overline\theta/\partial p<0$, poleward heat transport gives $F_p>0$, consistent with $F_p\propto\overline{v'T'}$. 
The representations use $F_\phi/(a\cos\phi)$, $F_p/(a\cos\phi)$, and $D=(a\cos\phi)^{-1}\nabla\!\cdot\!\mathbf F$, the implied acceleration of the zonal-mean zonal wind in m\,s$^{-1}$\,day$^{-1}$, which is undefined at the pole.
A non-zero divergence indicates that wave activity is deposited or removed and therefore transfers momentum to the zonal-mean wind. 
During an SSW, upward-propagating waves converge in the polar stratosphere, decelerating and potentially reversing the westerly polar-vortex winds.

\subsection{Northern Annular Mode}
\label{app:nam-convention}

EP flux diagnoses how waves propagate and force the zonal-mean circulation. We use the Northern Annular Mode (NAM) to measure the large-scale circulation state that accompanies this forcing across the atmospheric column \citep{baldwin_stratospheric_2001}.

Geopotential height \(Z(\lambda,\phi,p,t)\) is the physical height of a constant-pressure surface. We define its anomaly as
\[
Z'(\lambda,\phi,p,t)
=
Z(\lambda,\phi,p,t)
-
Z_{\mathrm{clim}}(\lambda,\phi,p,d(t)),
\]
where \(Z_{\mathrm{clim}}\) is the seasonally varying reference geopotential height and \(d(t)\) is the calendar day corresponding to time \(t\). Thus, \(Z'>0\) means that the pressure surface lies higher than expected for that location and time of year, while \(Z'<0\) means that it lies lower. 

At each pressure level \(p\), we identify the NAM pattern using principal component analysis. The resulting spatial component, \(e_p(\lambda,\phi)\), is conventionally called an empirical orthogonal function (EOF) in atmospheric science. We then measure how strongly the geopotential-height anomaly at time \(t\) resembles this pattern through the projection
\[
c(p,t)
=
\left\langle
\sqrt{\cos\phi}\,
Z'(\cdot,\cdot,p,t),
e_p
\right\rangle.
\]
Here, \(\langle\cdot,\cdot\rangle\) is a dot product over the latitude--longitude grid, and \(\sqrt{\cos\phi}\) accounts for the smaller area represented by grid points closer to the pole. A large positive projection means that the anomaly resembles \(e_p\); a large negative projection means that it resembles the opposite circulation pattern.

The dimensionless NAM index is the standardized projection,
\[
N(p,t)
=
\frac{c(p,t)-\mu_{c,p}}{\sigma_{c,p}},
\]
where \(\mu_{c,p}\) and \(\sigma_{c,p}\) are the reference mean and standard deviation at pressure level \(p\). The sign convention is chosen so that positive NAM represents a strong polar vortex with anomalously low polar geopotential height, while negative NAM represents a weakened or disrupted vortex with anomalously high polar geopotential height. 

The reference is fitted separately at each of the 37 ERA5 pressure levels on a $2.5^\circ$ grid between $20^\circ$ and $90^\circ$N, obtained by averaging daily-mean (00--18\,UTC) geopotential height over blocks of $10\times10$ native grid points, using 1979--2023 with 29 February excluded.
$Z_{\mathrm{clim}}$ is the calendar-day mean, low-pass filtered with a 90-day second-order Butterworth filter applied forward and backward.
$e_p$ is the leading EOF of the $\sqrt{\cos\phi}$-weighted, 90-day low-pass-filtered anomalies of November--April days, and $\mu_{c,p}$ and $\sigma_{c,p}$ are the mean and standard deviation of the projections of the unfiltered anomalies over all days of 1979--2023. The sign of $e_p$ is chosen so that positive NAM corresponds to negative polar-cap ($\geq60^\circ$N) height anomalies.
The index for 2024--2025 is obtained by projecting onto the same reference, and $N_{50:100}$ is the mean of the 50, 70, and 100\,hPa indices.

Computing \(N(p,t)\) at every pressure level and time produces a pressure--time section of the annular circulation state. Following an SSW, negative NAM values often first occur at low pressure (high altitude) in the stratosphere and subsequently appear at progressively higher pressure, corresponding to lower altitude. 

\section{Relationship Discovery and Selection}
\label{app:relationship-search}

This appendix specifies the relationship-discovery procedure summarized in
Section~\ref{sec:instantiating-expert-representations}.

\subsection{Data and cohort}

Field diagnostics are computed from the ARCO-ERA5 archive at $0.25^\circ$ resolution on 13 pressure levels (50--1000\,hPa) at 00, 06, 12, and 18\,UTC, and averaged over twelve $5^\circ$ latitude bands between $30^\circ$ and $90^\circ$N. They are standardized separately for each calendar day and synoptic hour with a 1979--2017 climatology, whose mean is smoothed with a 31-day running mean and whose standard deviation is not smoothed. The NAM uses all 37 pressure levels (Appendix~\ref{app:nam-convention}). The field diagnostics are extracted for November--April of 1979--2025.

An initialization $t_c$ (00\,UTC on a day from November to March) is admissible when its complete 40-day precursor history ($t_c-40$\,d to $t_c-6$\,h) lies within the extracted months and its subsequent 10-day NAM trajectory is available. Admissible initializations therefore fall between 11 December and 31 March. Winters are indexed by the year of their January, giving 46 complete winters from 1980 to 2025 and 5{,}118 admissible dates.
Restricting these dates to $|N_{50:100}(t_c)|\leq0.3$ yields 772 initializations (15.1\%), with 2 to 47 per winter (mean 16.8).

\subsection{Candidate source space}

Field representations are aggregated over five latitude regions,
\[
30\!-\!60^\circ\mathrm{N},\quad
38\!-\!68^\circ\mathrm{N},\quad
45\!-\!75^\circ\mathrm{N},\quad
52\!-\!82^\circ\mathrm{N},\quad
60\!-\!90^\circ\mathrm{N},
\]
using the constituent 5$^\circ$ latitude bands falling within each region.
For all non-divergence field representations, the four pressure regions are
\[
\{50,100,150\},\quad
\{200,250,300\},\quad
\{400,500\},\quad
\{600,700,850\}\ {\rm hPa},
\]
while EP-divergence representations use $\{600,700\}$\,hPa in the lowest region. Past NAM uses five pressure regions defined by the edges
\(
\{1,\ 50,\ 150,\ 400,\ 700,\ 1000\} \ {\rm hPa}.
\)
Temporal windows are
\(
W\in\{2,3,5,7,10,20,30\}\ {\rm days},
\)
and the five operators are
\[
\Psi=
\{\mathrm{mean},\mathrm{maximum},\mathrm{minimum},
\mathrm{positive\ impulse},\mathrm{negative\ impulse}\}.
\]
For samples $z_s$ at sampling rate $\nu$ per day, the impulse operators are
\[
\psi_W^+(z)=\frac{1}{\nu}\sum_{s\in W}\max(z_s,0),
\qquad
\psi_W^-(z)=\frac{1}{\nu}\sum_{s\in W}\max(-z_s,0).
\]
All windows terminate strictly before $t_c$. Combining 11 field representations, five latitude regions, four pressure regions, seven windows, and five operators gives 7,700 field candidates; the five NAM pressure regions contribute a further 175, for a total of 7,875.

\subsection{Split criterion and eligibility}
For each candidate source variable $X$, we search exhaustively over all threshold splits of the initializations being split: the neutral cohort $\mathcal T$ at the first level, and a fixed first-level branch at the second; below, $\mathcal T$ denotes this set in either case.
Let $x_{(1)}\leq\cdots\leq x_{(n)}$ denote the sorted source values. The candidate threshold set is
\begin{equation}
    \Gamma(X)
    =
    \left\{
        \frac{x_{(i)}+x_{(i+1)}}{2}
        :
        x_{(i)} < x_{(i+1)}
    \right\},
\end{equation}
so thresholds are placed only between distinct adjacent source values. Each $c\in\Gamma(X)$ partitions the initialization dates into two leaves,
\begin{equation}
    \mathcal T_0(c)=\{i:X_i\leq c\},
    \qquad
    \mathcal T_1(c)=\{i:X_i>c\}.
\end{equation}
Each initialization $i\in\mathcal T$ carries a weight $w_i\propto 1/n_{\mathcal T}(\omega_i)$, normalized so that $\sum_{i\in\mathcal T}w_i=1$, where $\omega_i$ is its winter and $n_{\mathcal T}(\omega)$ is the number of dates of $\mathcal T$ in winter $\omega$. Every winter therefore contributes the same total weight, however many of its dates enter the cohort (between 2 and 47).
For $\mathcal T'\subseteq\mathcal T$, let $\eta(\mathcal T')=\sum_{i\in\mathcal T'}w_i$ denote its summed weight and $\mu_\tau(\mathcal T')=\eta(\mathcal T')^{-1}\sum_{i\in\mathcal T'}w_iY_{i,\tau}$ its weighted mean NAM at lead day $\tau$, and let
\begin{equation}
    \mathcal V(\mathcal T')
    =
    \frac{1}{5}\sum_{\tau=6}^{10}
    \frac{1}{\eta(\mathcal T')}\sum_{i\in\mathcal T'}w_i\bigl(Y_{i,\tau}-\mu_\tau(\mathcal T')\bigr)^2
\end{equation}
be the weighted variance of the future 50--100\,hPa NAM, averaged over the scoring period (lead days 6--10).
The fraction of trajectory variance explained (FVE) by threshold $c$ is
\begin{equation}
    \operatorname{FVE}(c)
    =
    1-
    \frac{
        \sum_{b\in\{0,1\}}
        \frac{\eta(\mathcal T_b(c))}{\eta(\mathcal T)}
        \mathcal{V}\!\left(\mathcal T_b(c)\right)
    }{
        \mathcal{V}(\mathcal T)
    }.
\end{equation}
The weights are computed for the node being split: the full cohort at the first level, and the fixed branch at the second level. They depend only on the initialization dates, so they are identical in every permutation null universe. In the stability refits, which add or remove whole winters, each remaining winter also keeps equal weight.

The selected threshold maximizes FVE over the midpoints that leave at least $n_{\min}$ dates in each child,
\begin{equation}
    c^\star
    =
    \arg\max_{c\in\Gamma(X):\ |\mathcal T_0(c)|,\,|\mathcal T_1(c)|\,\geq\, n_{\min}}
    \operatorname{FVE}(c),
\end{equation}
with $n_{\min}=50$ at the first level and $n_{\min}=30$ at the second.
Of the two children of $c^\star$, the selected child $\mathcal T^\star$ is the one whose weighted day 6--10 mean 
\begin{equation}
\bar\mu(\mathcal T')=\frac15\sum_{\tau=6}^{10}\mu_\tau(\mathcal T')
\end{equation}
lies furthest from $\bar\mu(\mathcal T)$. Because
\begin{equation}
\eta(\mathcal T_0)\bar\mu(\mathcal T_0)+\eta(\mathcal T_1)\bar\mu(\mathcal T_1)=\eta(\mathcal T)\bar\mu(\mathcal T),
\end{equation}
this is the child with the smaller summed weight.

Two winter-coverage requirements prevent a high FVE from being obtained by isolating behaviour confined to one or a few winters. They are applied to the optimized split, and a candidate that fails them is discarded rather than refitted. At the first level, the selected child must span at least five distinct winters; both children of every retained first-level split then define branches for the second-level search. At the second level, the selected child must again span at least five winters, and both children must each span at least eight.

The split search is exhaustive: every eligible midpoint is evaluated, rather than selecting thresholds from a fixed grid or a predefined set of quantiles. Consequently, the reported FVE for a candidate source is the maximum trajectory separation attainable by a single threshold subject to the minimum child size.

\subsection{Winter-level stability filtering}
Maximizing FVE on the complete cohort can favor relationships whose apparent separation depends on the particular winters used for discovery. 
We therefore apply a second-stage stability filter that asks whether the same candidate source repeatedly produces a separation in the same direction when thresholds are estimated from one set of winters and evaluated on different winters.

We generate 1{,}000 repeated 70/30 train--held-out splits of the winters represented in the node (the 46 winters of the cohort at the first level, or those of the branch at the second), shared by all candidates in that node.
The resampling unit is the winter: all initialization dates belonging to a given winter are assigned jointly to either the training or held-out subset. 
This preserves the temporal dependence among initialization dates from the same winter and prevents densely sampled winters from being treated as collections of independent observations.

The candidate source specification itself is fixed throughout this procedure: its representation, spatial region, temporal window, and temporal operator are not reselected. For resample $s$, only its threshold is re-estimated. Using the training winters, we repeat the threshold search described above, with the same minimum child size and five-winter requirement on the selected child,
\begin{equation}
    c_s^\star
    =
    \arg\max_{c}
    \operatorname{FVE}_{\mathrm{train},s}(c).
\end{equation}
The resulting threshold $c_s^\star$ is then transferred unchanged to the held-out winters. Thus, the held-out data play no role in choosing either the candidate source variable or its threshold.

Let
\begin{equation}
d_{\mathrm{full}}=\operatorname{sign}\bigl(\bar\mu(\mathcal T^\star)-\bar\mu(\mathcal T)\bigr)
\end{equation}
be the direction in which the selected child of the full fit shifts the day 6--10 NAM.
In resample $s$, the training refit determines both $c_s^\star$ and which of its sides is selected. 
Applying this rule unchanged to the held-out dates $\mathcal T_s^{\mathrm{held}}$ gives their selected child $\mathcal T_s^{\mathrm{held},\star}$ and
\begin{equation}
d_s^{\mathrm{held}}=\operatorname{sign}\bigl(\bar\mu(\mathcal T_s^{\mathrm{held},\star})-\bar\mu(\mathcal T_s^{\mathrm{held}})\bigr),
\end{equation}
with weights recomputed on $\mathcal T_s^{\mathrm{held}}$.
A resample is valid when the training refit has an admissible split and both held-out children are non-empty. With $\mathcal S_{\mathrm{valid}}$ the set of valid resamples, we define directional stability as
\begin{equation}
    S_{\mathrm{dir}}
    =
    \frac{1}{|\mathcal S_{\mathrm{valid}}|}
    \sum_{s\in\mathcal S_{\mathrm{valid}}}
    \mathbb{I}
    \left[
        d_s^{\mathrm{held}}
        =
        d_{\mathrm{full}}
    \right].
\end{equation}

This criterion requires the direction of the relationship to reproduce in at least 95\% of the repeated held-out-winter evaluations, despite re-estimation of the threshold from different training winters. 
Importantly, the 95\% threshold is an operational reproducibility criterion, not a $p$-value, confidence level, or test of statistical significance. 
Its purpose is to remove relationships whose fitted effect is highly sensitive to which winters are available for estimation. Chance discovery and multiplicity across the large candidate catalogue are handled separately by the cross-winter permutation-FDR procedure described below.
All 283 retained first-level relationships have at least 999 valid resamples. Small second-level branches admit fewer valid refits: the median over the 672 retained second-level relationships is 998, but 61 have fewer than 500 and 23 fewer than 20.

\subsection{Permutation-FDR calibration}

Directional stability removes relationships that fail to reproduce across held-out winters, but it does not account for the multiplicity induced by searching a large catalogue of candidate relationships. Even under no genuine precursor--future association, an exhaustive search over thousands of representations, spatial regions, temporal windows, operators, and thresholds may produce apparently strong and stable relationships by chance. We therefore calibrate the full discovery procedure against $B=100$ permutation null universes.

Each null universe is constructed by reassigning complete future NAM trajectories across winters. Let $Y_i\in\mathbb{R}^{K}$ denote the full future NAM trajectory associated with initialization $i$. In permutation universe $b$, the precursor variables and initialization dates remain unchanged, while the target becomes
\begin{equation}
    Y_i^{(b)} = Y_{\pi_b(i)},
\end{equation}
where $\pi_b$ is constrained to assign trajectories from different winters.
Each $\pi_b$ is a random permutation of the 772 initializations in which no initialization receives a trajectory from its own winter; the second-level search uses the same 100 permutations, restricted to each branch.
The complete trajectory is reassigned as a single vector rather than permuting individual lead times. This preserves the temporal dependence within each future trajectory while breaking its association with the precursor conditions from which it originally evolved.

For every permutation universe, we repeat the same discovery procedure applied to the real ERA5 data: all candidate sources are searched, thresholds are optimized subject to the same eligibility criteria, and the same 1{,}000 winter-level stability procedure and $S_{\mathrm{dir}}\geq0.95$ criterion are applied. A permutation universe therefore represents one complete realization of the relationship-discovery pipeline under a null in which precursor and future trajectories are disconnected.

For an FVE threshold $t$, let
\begin{equation}
    R(t)
    =
    \#\left\{
        \text{real stability-passing relationships with FVE}\geq t
    \right\},
\end{equation}
and let
\begin{equation}
    N_b(t)
    =
    \#\left\{
        \text{stability-passing relationships in null universe }b
        \text{ with FVE}\geq t
    \right\}.
\end{equation}
The expected number of discoveries produced by the same search under the null is estimated by averaging across permutation universes,
\begin{equation}
    \overline{N}(t)
    =
    \frac{1}{B}
    \sum_{b=1}^{B} N_b(t),
\end{equation}
giving the empirical false-discovery-rate (FDR) estimate
\begin{equation}
    \widehat{\mathrm{FDR}}(t)
    =
    \frac{\overline{N}(t)}{R(t)}
    =
    \frac{
        B^{-1}\sum_{b=1}^{B}N_b(t)
    }{
        R(t)
    }.
\end{equation}
Thus, $\widehat{\mathrm{FDR}}(t)$ compares the number of relationships retained from the real catalogue with the number expected to survive the complete search-and-stability pipeline when the precursor--future association has been destroyed.

Because only $B=100$ null universes are available, we additionally quantify uncertainty in this estimate by bootstrapping the permutation universes. Each bootstrap replicate resamples the $B$ complete null universes with replacement, recomputes $\overline{N}(t)$, and hence recomputes $\widehat{\mathrm{FDR}}(t)$. We use the upper end of the 95\% bootstrap percentile interval, the 97.5th percentile over 20{,}000 bootstrap replicates, as a conservative calibration criterion and retain the smallest FVE threshold $t^\star$ satisfying
\begin{equation}
    q_{0.975}\!\left[
        \widehat{\mathrm{FDR}}(t^\star)
    \right]
    \leq 10^{-3},
\end{equation}
where $q_{0.975}$ denotes the 97.5th percentile over the bootstrap replicates.

The permutation universes, rather than individual null candidates, are the resampling units in this calibration: each universe represents one complete null realization of the entire catalogue search. The resulting threshold therefore bounds the estimated proportion of retained relationships that the full search produces under the cross-winter null, conditional on having already passed the winter-level reproducibility filter.

\subsection{First- and second-level discovery}

At the first level, each of the 7{,}875 candidate source variables is evaluated independently against the same future NAM target. For every source, the best eligible threshold is selected by maximizing FVE as described above, after which the relationship is subjected to the 1{,}000-resample winter-level stability procedure. Of the 7{,}875 candidates, 295 satisfy
\begin{equation}
    S_{\mathrm{dir}}\geq0.95.
\end{equation}
These stability-passing relationships are then calibrated against the first-level permutation null. The smallest FVE threshold whose upper bootstrap bound on the empirical FDR (the 97.5th percentile over 20{,}000 replicates) is at most $10^{-3}$ is approximately
\begin{equation}
    t_1^\star \simeq 0.0504,
\end{equation}
leaving 283 retained first-level relationships.

Each retained first-level relationship consists of a source specification, an optimized threshold, and the resulting partition of the 772 initialization dates into two child leaves. These quantities are frozen after first-level selection. Since every retained split defines two branches, the 283 relationships produce
\begin{equation}
    2\times 283 = 566
\end{equation}
fixed first-level branches.

Second-level discovery asks whether additional precursor information distinguishes future NAM evolution conditional on membership in one of these first-level branches. Within each of the 566 fixed branches, we therefore search exhaustively over all second-source variables from the same 7{,}875-variable catalogue, subject to the second-level eligibility requirements. The first-level source, threshold, and branch membership remain fixed throughout; only the second-level source and its threshold are optimized.

Across all branches, this exhaustive conditional search evaluates 4{,}358{,}681 eligible second-level candidate relationships. Each candidate is then subjected to the same winter-level threshold re-estimation and directional stability procedure used at the first level. Of these candidates, 71{,}494 satisfy $S_{\mathrm{dir}}\geq0.95$.

The second-level permutation analysis is conditional on the first-level structure discovered in ERA5. 
In every null universe, we therefore retain exactly the same 283 first-level relationships, their thresholds, and the resulting 566 branch memberships. The first-level search is not repeated under the null. 
Future NAM trajectories are permuted as described above, and the complete second-level search and stability filtering are rerun within these fixed branches. 
The resulting null therefore asks whether apparent additional predictive structure of comparable strength could arise by chance once the first-level partition has already been fixed.

Applying the same permutation-FDR criterion to the second-level search gives a conditional-FVE threshold of approximately
\begin{equation}
    t_2^\star \simeq 0.3102,
\end{equation}
retaining 672 second-level relationships. These relationships define the conditional predictive structures used for subsequent model evaluation.

\subsection{Retained relationship panel}

The 672 statistically retained second-level relationships form the primary evaluation panel. Discovery is completed entirely from the reference ERA5 trajectories before any forecast model is evaluated. For each retained relationship, the first-level source and threshold, first-level branch, second-level source and threshold, selected terminal leaf, and initialization dates belonging to that leaf are therefore frozen.

When evaluating GraphCast, FengWu, and Pangu-Weather, no representation, spatial region, temporal window, operator, threshold, branch assignment, or case membership is re-estimated from the model forecasts. Each model is instead evaluated on the same ERA5-defined conditional subset. 
Differences between the reference and model-generated future NAM distributions therefore measure whether the model reproduces the consequence of a relationship identified independently in the reference system, rather than whether the model can define an alternative partition that better fits its own dynamics.

The exhaustive catalogue nevertheless produces some closely related retained relationships. 
For example, two relationships may describe the same underlying configuration while differing only in latitude or pressure region, temporal window, or aggregation operator. 
To assess sensitivity to this redundancy, we group the retained second-level relationships within each fixed first-level branch. 
Two relationships in the same branch are linked when their second-level sources share a representation and differ in exactly one of latitude region, pressure region, temporal window, or operator. 
Motifs are the connected components of these links, so relationships joined by a chain of such single-dimension differences share a motif, and relationships in different branches are never grouped. This rule reduces the 672 retained relationships to 287 motifs; 179 contain a single relationship, and the largest contains 39.

The 672 raw relationships remain the primary evaluation panel because they are the direct output of the statistical selection procedure. As a sensitivity analysis, we first average results among relationships belonging to the same motif and then weight the resulting 287 motifs equally. This prevents parent branches containing many closely related catalogue variants from receiving disproportionate weight. The collapse should be interpreted only as a catalogue-level redundancy correction: the resulting 287 motifs are not assumed to be statistically independent or to correspond to 287 distinct physical mechanisms.

Table~\ref{tab:motif-sensitivity} compares the two weightings over lead days 6--10. The model ranking is unchanged under both: Pangu-Weather has the lowest conditional-mean and conditional-variance errors, and FengWu the lowest $T$, $u$, and $v$ MAE; the conventional errors change by at most 0.02.
The main difference is FengWu's conditional-mean error, which rises from 0.148 to 0.169 and nearly reaches GraphCast's 0.171.
FengWu's relationship-level advantage over GraphCast is concentrated in the ten largest motifs. These hold 191 relationships, 180 of them in branches of 600--850\,hPa zonal-mean $u$ at 38--68$^\circ$N, and there FengWu's conditional-mean error is 0.10 against GraphCast's 0.16. Among the 179 single-relationship motifs, GraphCast has the lower error (0.16 against 0.18).

\begin{table}[h]
\centering
\small
\caption{Lead-day 6--10 errors averaged over the 672 retained relationships and over the 287 motifs (relationships averaged within each motif, motifs weighted equally). $T$, $u$, and $v$ are forecast MAE in K and m\,s$^{-1}$.}
\label{tab:motif-sensitivity}
\begin{tabular}{llccccc}
\toprule
Model & Average over & $E_\mu$ & $E_{\sigma^2}$ & $T$ & $u$ & $v$ \\
\midrule
\multirow{2}{*}{GraphCast}     & 672 relationships & 0.173 & 0.195 & 2.290 & 5.729 & 5.768 \\
                               & 287 motifs        & 0.171 & 0.179 & 2.285 & 5.717 & 5.748 \\
\midrule
\multirow{2}{*}{FengWu}        & 672 relationships & 0.148 & 0.210 & 1.762 & 4.488 & 4.558 \\
                               & 287 motifs        & 0.169 & 0.205 & 1.765 & 4.489 & 4.556 \\
\midrule
\multirow{2}{*}{Pangu-Weather} & 672 relationships & 0.091 & 0.107 & 2.175 & 5.494 & 5.581 \\
                               & 287 motifs        & 0.090 & 0.104 & 2.180 & 5.498 & 5.579 \\
\bottomrule
\end{tabular}
\end{table}

\section{Machine learning weather models}
\label{app:ml_models}

We evaluate three published, pretrained machine-learning weather forecasting systems: \textbf{GraphCast} (operational checkpoint), \textbf{Pangu-Weather} (6-hour configuration), and \textbf{FengWu} (operational checkpoint). 
All forecasts are generated through NVIDIA's \texttt{earth2studio} (v0.13.0) inference framework and initialized from the same ERA5 source. The models retain their native input requirements and inference procedures, but are evaluated using a common forecast horizon, output cadence, variable set, and verification pipeline.

\subsection{GraphCast (operational)}

GraphCast is a graph-neural-network weather forecasting model that represents global atmospheric interactions using an internal multimesh graph while ingesting and producing fields on a regular latitude--longitude grid~\citep{lam_learning_2023}. 
We use the published operational checkpoint at $0.25^\circ$ horizontal resolution with 13 pressure levels and a native 6-hour forecast step. 
This checkpoint was pretrained on ERA5 over 1979--2017 and subsequently fine-tuned on operational IFS HRES analyses over 2016--2021.

\subsection{Pangu-Weather (6-hour)}

Pangu-Weather is a three-dimensional Earth-Specific Transformer designed for global medium-range weather prediction~\citep{bi_accurate_2023}.
The published system comprises networks associated with different forecast intervals. 
We use the \texttt{earth2studio} 6-hour configuration, which combines the published 24-hour and 6-hour Pangu-Weather networks to produce forecasts at 6-hour intervals throughout the rollout. 
The model operates at $0.25^\circ$ horizontal resolution on 13 pressure levels.
The published Pangu-Weather models were trained using ERA5 data from 1979--2017.

\subsection{FengWu (operational)}

FengWu is a multimodal, multitask transformer for global medium-range weather forecasting~\citep{chen_operational_2025}.
We use the \texttt{fengwu\_v1} checkpoint released by the authors and distributed through \texttt{earth2studio}: a single autoregressive model with a native 6-hour forecast step, operating at $0.25^\circ$ resolution on 13 pressure levels with 69 input variables.
It was trained on ERA5 over 1979--2017 and does not include the subsequent transfer learning on ECMWF operational analyses (2017--2021) used for the operationally deployed FengWu.

\subsection{Common forecast protocol}

All three models are initialized from the Analysis-Ready, Cloud-Optimized (ARCO) ERA5 archive, which provides hourly fields at $0.25^\circ$ resolution on 37 pressure levels. Forecasts are initialized independently for each of the 772 neutral-NAM dates defined in Section~\ref{sec:nam_target}.
Only variables and pressure levels required by the respective native model inputs are extracted from ERA5.

Each initialization is propagated autoregressively for 11 days ($264$\,h), producing forecasts at
\[
\{0,6,12,\ldots,264\} \ \mathrm{h}.
\]
The resulting 6-hourly outputs are subsequently grouped by forecast day for the analyses reported in this work.

Evaluation uses the 13 pressure levels shared by all three forecast systems,
\[
\{50,\ 100,\ 150,\ 200,\ 250,\ 300,\ 400,\ 500,\ 600,\ 700,\ 850,\ 925,
\ 1000\} \ \mathrm{hPa},
\]
and the implementation checks that every model outputs $T$, $u$, $v$, and $z$ on all of them. Forecast lead day $d$ is the calendar day $t_c+d$, i.e.\ the four forecasts valid at 00, 06, 12, and 18\,UTC on that day; lead days 1--10 are evaluated.

Two quantities are computed from every rollout.
First, conventional forecast errors of $T$, $u$, and $v$ are evaluated against a $1^\circ$ ERA5 verification archive. Model output on the native $0.25^\circ$ grid is subsampled to every fourth grid point, so verification points coincide with native model grid points rather than being interpolated. The daily MAE is the unweighted mean absolute error over all $1^\circ$ grid points (without area weighting), the 13 pressure levels, and the four 6-hourly forecasts of the day. Within a terminal leaf it is averaged over initializations with the same winter weights as $\mu_{m,r}$ and $\sigma^2_{m,r}$ (Section~\ref{sec:relationship-search}).

Second, the forecast NAM is obtained by averaging the four 6-hourly geopotential fields of each day, converting to height, averaging to the $2.5^\circ$ reference grid over blocks of $10\times10$ grid points, and projecting onto the ERA5 reference EOFs using the ERA5 climatology and normalization (Appendix~\ref{app:nam-convention}).
Because 70\,hPa is not among the models' output levels, the model $N_{50:100}$ is the mean of the 50 and 100\,hPa indices, whereas the ERA5 reference averages 50, 70, and 100\,hPa. Using the 50/100\,hPa mean for ERA5 as well changes the lead-day 6--10 $E_\mu$ and $E_{\sigma^2}$ of Table~\ref{tab:representation-model-evaluation} by at most 0.004 and 0.007, respectively, and leaves the model ranking unchanged.
The NAM reference contains no 29 February. Model lead days falling on that date therefore have no NAM value and are omitted from that day's statistics, while the ERA5 target on 29 February is interpolated from the neighbouring days.

All precursor sources are computed from ERA5: every relationship is defined by ERA5 histories before $t_c$, and the models enter the evaluation only through their NAM forecasts and forecast errors.

\subsection{Forecast coverage}

All $772\times3=2{,}316$ model--initialization rollouts are complete for lead days 1--10, and no forecast value is masked, clipped, or imputed. The largest absolute forecast $N_{50:100}$ over the cohort is 3.51 for GraphCast, 3.71 for FengWu, and 3.89 for Pangu-Weather.

\subsection{NAM forecast error}
To separate conditional fidelity from forecast accuracy on the target variable itself, we also compute the plain NAM forecast error, the winter-weighted mean absolute difference between forecast and ERA5 $N_{50:100}$ over lead days 6--10. 
Over all 772 initializations it is 0.453 for GraphCast, 0.294 for FengWu, and 0.275 for Pangu-Weather (0.494, 0.337, and 0.290 averaged over the 672 terminal leaves). 
Pangu-Weather therefore also has the lowest NAM forecast error, but its advantage over FengWu is far smaller than in conditional-mean error (0.091 against 0.148) or conditional-variance error (0.107 against 0.210). 
NAM forecast error also does not reproduce the relationship-level differences: FengWu has a lower NAM forecast error than GraphCast in every row of Table~\ref{tab:representation-model-evaluation}, yet GraphCast has the lower conditional-mean error in five of them, including strong negative shifts (NAM forecast error 0.590 against 0.456; conditional-mean error 0.275 against 0.329). 
GraphCast also has the smallest unconditional mean error over all initializations (0.010, against 0.067 for FengWu and 0.032 for Pangu-Weather) but the largest conditional-mean error, so its errors appear only once the initializations are conditioned on the discovered precursors. Using the 50/100\,hPa mean for ERA5 changes these values by at most 0.004.
This is an empirical counterpart of Appendix~\ref{app:coarsening}: agreement in aggregate statistics does not imply agreement in the conditional evolution.

To give these errors a scale, we score two reference forecasts that carry no information about the precursors in the same way: climatology, which predicts the cohort-mean ERA5 trajectory for every initialization, and persistence of the initial ERA5 $N_{50:100}$.
Their NAM forecast errors over all initializations are 0.954 and 0.937, and averaged over the 672 terminal leaves their conditional-mean and conditional-variance errors are 0.714 and 0.668, and 1.172 and 1.144, several times those of all three models.
The exceptions are the weak-shift rows of Table~\ref{tab:representation-model-evaluation}, whose leaves by construction remain close to the cohort mean: there the references reach conditional-mean errors comparable to the models' (for weak negative shifts, 0.099 for climatology and 0.075 for persistence against 0.078--0.128), while their conditional-variance errors remain 5--10 times larger, so these rows mainly test the conditional variance.

\subsection{Training-period overlap with the evaluation cohort}

The reference cohort spans winters from 1979--2025 and therefore overlaps the training periods of the published forecast systems. 
GraphCast's operational checkpoint was pretrained on ERA5 over 1979--2017 and subsequently fine-tuned on IFS HRES operational analyses over 2016--2021.
The published Pangu-Weather and FengWu models were trained on ERA5 over 1979--2017.
Thus, a substantial fraction of the 772 initialization dates lies within periods used in the development or training of these models, whereas later years provide temporally out-of-sample cases. Of the 772 initializations, 648 fall in 2017 or earlier, within the ERA5 training period of all three models; 66 fall in 2018--2021, within GraphCast's HRES fine-tuning period; and 58 fall in 2022--2025.

The purpose of the present experiment is therefore not to construct a strictly held-out benchmark of general forecast skill. 
Instead, all three models are evaluated on the same reference-defined dynamical relationships across the full available cohort. 
The overlap should nevertheless be kept in mind when interpreting absolute forecast performance. 

\end{document}